\documentclass{article}

\PassOptionsToPackage{numbers, compress}{natbib}

 \usepackage[main, final]{neurips_2026}

\usepackage[utf8]{inputenc} %
\usepackage[T1]{fontenc}    %
\usepackage{hyperref}       %
\usepackage{url}            %
\usepackage{booktabs}       %
\usepackage{amsfonts}       %
\usepackage{nicefrac}       %
\usepackage{microtype}      %
\usepackage{xcolor}         %
\usepackage{amsmath}
\usepackage{amssymb}
\usepackage{nicefrac}
\usepackage{microtype}
\usepackage{graphicx}
\usepackage{xcolor}
\usepackage{algorithm}
\usepackage{algorithmic}
\usepackage{natbib}
\usepackage{multirow}
\usepackage{caption}
\usepackage{subcaption}
\usepackage{float}
\usepackage{eqparbox}
\usepackage{paralist}
\usepackage{enumitem}
\usepackage{xspace}

\usepackage[capitalize]{cleveref}

\newcommand{\codename}{\textsc{EvoLib}\xspace}

\title{Test-Time Learning with an Evolving Library}

\author{%
  Weijia Xu\textsuperscript{1} \And
  Alessandro Sordoni\textsuperscript{1} \And
  Chandan Singh\textsuperscript{1} \And
  Zelalem Gero\textsuperscript{1} \AND
  Michel Galley\textsuperscript{1} \And
  Xingdi Yuan\textsuperscript{1} \And
  Jianfeng Gao\textsuperscript{1} \AND\\
  \textsuperscript{1}Microsoft Research \\
  Correspondence: \texttt{weijiaxu@microsoft.com}
}

\begin{document}

\maketitle

\begin{abstract}
We introduce \codename, a test-time learning framework that enables large language models to accumulate, reuse, and evolve knowledge across problem instances without parameter updates or external supervision. 
Instead of adapting model parameters, our approach maintains a shared library of \emph{knowledge abstractions}, including modular skills and reflective insights, automatically extracted from the model's own inference trajectories. 
To support continual improvement, we introduce a principled weighting and consolidation mechanism that jointly optimizes for immediate utility and long-term value. This allows simple, instance-specific abstractions to evolve into more general and reusable ones over time. 
Across challenging benchmarks in mathematical reasoning, code generation, and multi-turn agentic environments, \codename improves substantially over the top test-time scaling and learning methods without ground-truth feedback.\footnote{Code is available at: \url{https://github.com/microsoft/EvoLib}.}
\end{abstract}

\section{Introduction}
\label{sec:intro}

Humans can learn from past experience and encode that knowledge in abstract, reusable, and extensible forms. This allows them to accumulate and refine knowledge over time, enabling adaptation to novel situations with far less repeated effort. By contrast, large language models~(LLMs) typically handle each input problem independently~\citep{wang2023selfconsistency,weng2023selfverification,snell2024scaling,zhuang2026test,muennighoff2025s1, venkatraman2025rsa} or share memory across problem instances by storing and retrieving past problems and trajectories from a global memory~\citep{madaan2022memory,feng2024thoughtretriever,wei2025evomemory}, which hinders LLMs from inducing generalized knowledge or skills across problems.

\emph{Test-time learning}~(TTL) bridges this gap by enabling models to adapt during inference, which improves performance across a sequence of tasks~\cite{hu2025testtime,akyurek2025tttfewshot,yuksekgonul2026learning}. In principle, TTL allows models to correct systematic errors, discover reusable strategies, and build knowledge across problems. However, existing approaches face two fundamental limitations in the context of modern LLMs. 
First, most methods rely on gradient updates on model parameters to learn effectively from test-time signals. These approaches are not applicable in the increasingly common \emph{black-box} setting, where model weights are inaccessible. Second, many TTL approaches depend on external supervision signals, such as reward functions or example test cases, to guide learning. Such signals are unavailable in many real-world application scenarios.

A complementary line of work, \emph{test-time scaling}, improves performance for black-box LLMs without external supervision, by allocating more computation per instance~(e.g. through sampling~\cite{wang2023selfconsistency}, searching~\cite{venkatraman2025rsa}, or iterative refinement~\cite{zhuang2026test}). While effective, these methods treat each problem in isolation: the knowledge uncovered during the search process is discarded after inference, preventing the model from inducing generalizable knowledge across problems. 

\looseness=-1
In this paper, we introduce \codename, an evolving library for test-time learning that overcomes these limitations by enabling \emph{knowledge accumulation without parameter updates or external supervision}. The key idea is to maintain a structured, evolving collection of \emph{knowledge abstractions}. The system extracts two types of abstractions from model-generated solutions: (1) \emph{modular skills}, which capture reusable procedures, and (2) \emph{reflective insights}, which encode common errors and corrective strategies.

\codename is governed by two central principles:
\begin{itemize}[leftmargin=*]
    \item \textbf{Abstraction:} Instead of memorizing raw experiences, the library distills them into reusable knowledge units that can be flexibly recomposed across tasks. This enables more efficient transfer compared to memory-based approaches that store unstructured trajectories.
    \item \textbf{Evolution:} The library is continuously updated through iterative abstraction extraction, consolidation and dynamic weight computation. We introduce a novel credit assignment mechanism based on \emph{Information Gain} (IG) and \emph{Future Information Gain} (Future IG), which jointly capture the immediate usefulness of an abstraction and its potential for generating valuable future abstractions. This mechanism promotes the emergence of increasingly useful and extensible abstractions over time.
\end{itemize}
Importantly, the entire process is \emph{self-supervised} \---\ the model evaluates its own generated solutions and extracts abstractions without relying on external ground-truth signals. This makes the approach applicable to highly challenging tasks and deployment scenarios where external feedback is unavailable or expensive to collect.

We evaluate \codename on a diverse suite of benchmarks spanning mathematical reasoning, code generation, and multi-turn agentic tasks. Under matched compute budgets, \codename outperforms strong test-time scaling and test-time learning methods including Dynamic Cheatsheet~\citep{suzgun2025dc} and ReasoningBank~\citep{ouyang2026reasoningbank}, and improves more rapidly as compute increases. Its gains persist when the self-evaluation signal is noisy and when a weaker judge is used. In the continual learning setting, \codename is also robust to task ordering, with the largest advantage over existing TTL methods when heterogeneous tasks are randomly interleaved.

\section{Related Work}
\label{sec:related}

\paragraph{Test-time scaling and adaptation.}
There is a growing body of work that improves LLM performance by allocating additional computation at inference time rather than updating model parameters. Representative approaches include best-of-$N$ sampling and self-consistency~\citep{wang2023selfconsistency}, self-verification with and without feedback~\citep{weng2023selfverification,snell2024scaling,zhuang2026test}, iterative refinement and aggregation~\citep{muennighoff2025s1, venkatraman2025rsa}. While effective, these methods treat each problem independently: any knowledge discovered during inference is discarded after solving a single instance. By contrast, \codename explicitly retains and reuses knowledge across problems, enabling continual improvement over a stream of tasks.

More closely related are \emph{test-time learning} and \emph{test-time training} methods that adapt model behavior at test time. Recent work formulates test-time learning for LLMs as self-supervised optimization on unlabeled test data~\cite{hu2025testtime} or few-shot learning on example test cases~\cite{akyurek2025tttfewshot}, often using lightweight parameter updates~\cite{sun2024learning,kuratov2026gradmem}.
These methods require gradient updates on model parameters, whereas \codename performs test-time learning purely through abstraction induction and evolution, without any parameter modification.

\paragraph{Memory-augmented LLMs.}
Several recent works augment LLMs with external memory to persist information across queries or interactions. Retrieval-augmented generation and dynamic prompt construction store and retrieve raw past experience~\citep{madaan2022memory,feng2024thoughtretriever}.
\cite{wei2025evomemory} introduces ExpRAG, a simple framework for retrieving and reusing prior experience, along with a variant that allows the agent to iteratively refine the retrieved memory.
While these approaches demonstrate the importance of shared memory across queries, they primarily store unstructured trajectories or episodic experiences. \codename differs by distilling experience into \emph{extensible abstractions} and by explicitly promoting extension and consolidation to produce more general knowledge units.

\paragraph{Skill and knowledge discovery.}
The idea of learning reusable skills from experience has a long history in program synthesis and cognitive science~\citep{tenenbaum2011grow,vanlehn1996cognitive,chollet2019measure}. One line of work learns how to use existing tools effectively~\citep{prabhu2025walt,lu2026skill0}. Another line of work focuses on extracting reusable skills, such as subroutines or functions, from solved programs or successful trajectories~\citep{stengeleskin2024regal, wang2025inducing, liu2026reuseit, qiu2026autorefine,ho2025arcmemo}, or inducing reflective insights from model-generated programs or trajectories together with feedback~\citep{zhao2024expel, wu2025evolver, shi2026evolving,qu2025rlad,xie2026hybrid}. Furthermore, inspired by the recent development of agentic ``autoresearch'' loops~\cite{lu2024ai,novikov2025alphaevolve,wang2025thetaevolve}, recent works propose skill evolution by iteratively refining a skill set based on new trajectories~\cite{ma2026skillclaw,alzubi2026evoskill}.
Most existing approaches learn such libraries on training tasks where external success signals are available, whereas our approach can be applied solely at test-time.

Closest to our setting, AWM~\cite{wang2025awm} proposes to maintain a memory of reusable routines extracted from action trajectories with optional external feedback, while
Dynamic Cheatsheet~\citep{suzgun2025dc} induces high-level principles and procedural knowledge via self-reflection. Similarly, ReasoningBank constructs a memory of reasoning strategies distilled from self-judged successful and failed experiences~\cite{ouyang2026reasoningbank}.
However, these approaches either treat memory as a monolithic artifact that is refined holistically~\cite{suzgun2025dc} or append new items into the memory in a linear fashion~\cite{wang2025awm,ouyang2026reasoningbank}, making them susceptible to three major failure modes: \emph{misgrouping}, where experiences from different types of tasks are consolidated into the same abstraction; \emph{interference}, where repeated LLM rewriting strips away applicability conditions and turns useful lessons into overly broad guidance; and \emph{overfit}, where memory becomes tied to narrow surface patterns rather than reusable strategies~\cite{zhang2026usefulmemoriesfaultycontinuously}. 
\codename addresses these issues by introducing a principled weighting, sampling, and consolidation mechanism that drives the emergence and nonlinear evolution of abstractions, guided by immediate and long-term utility across instances (see the extended results in \cref{sec:extended_results}).

\section{\codename}
\label{sec:method}

\begin{figure}[t]
    \centering
    \includegraphics[width=.9\textwidth]{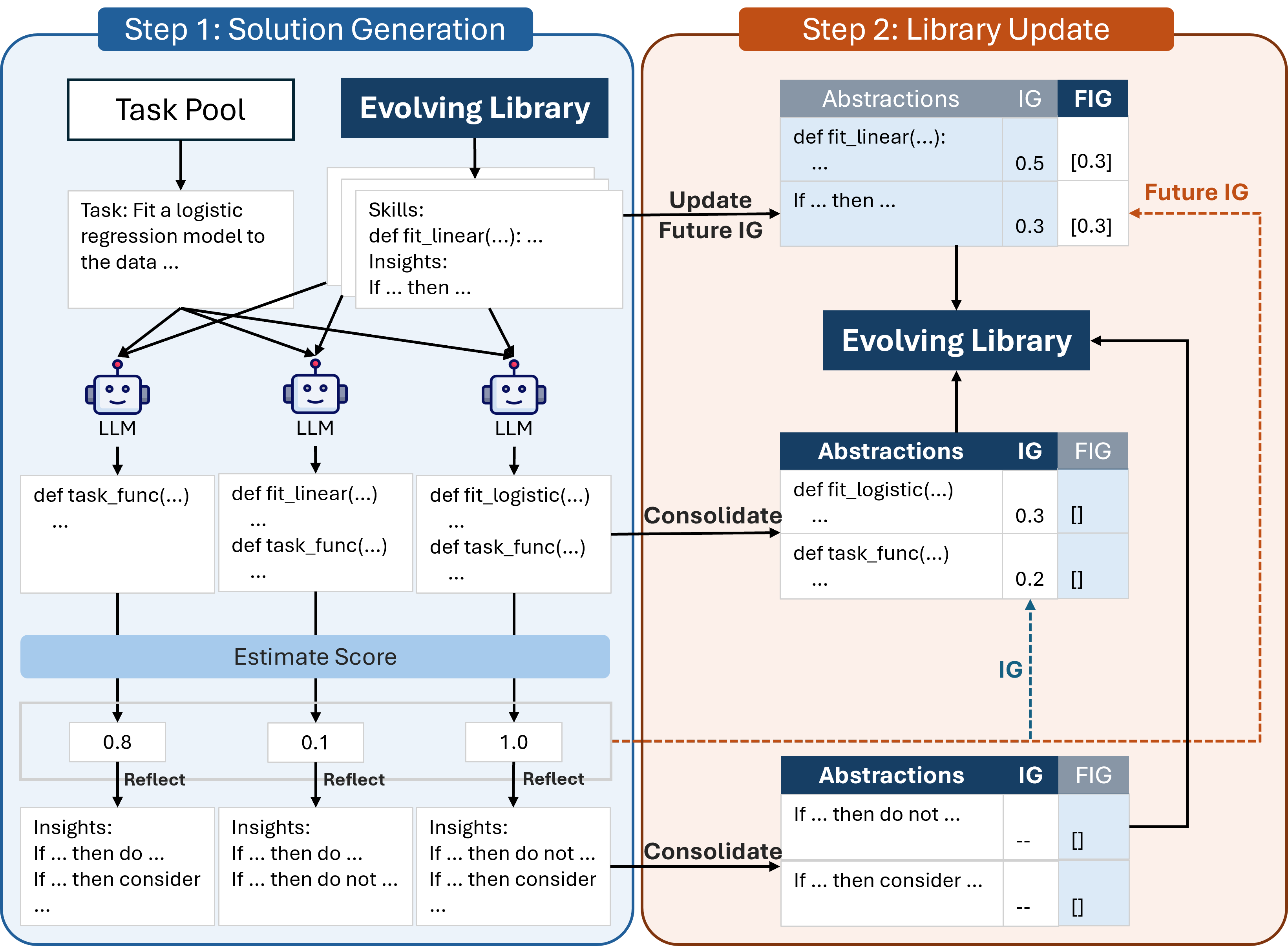}
\caption{Overview of the \codename algorithm. \codename performs test-time learning by repeatedly: (i) solving tasks using sampled abstractions, (ii) extracting new abstractions, consolidating them into the library, and propagating credit to both new and previously used abstractions via Information Gain~(IG) and Future IG.}
\label{fig:algorithm}
\end{figure}

\codename is characterized by two key properties. First, \textbf{abstraction} \---\ the library stores generalizable knowledge distilled from raw trajectories, making it easier for LLMs to adapt and reuse past experience across problems compared to memory-based methods that store raw trajectories. Second, \textbf{evolution} \---\ the library is continuously updated through a consolidation mechanism and a dynamic weighting scheme based on Information Gain and Future Information Gain, enabling abstractions to improve over time.

\subsection{Problem Formulation}
\label{sec:problem_formulation}
We represent an agent augmented with \codename as a tuple~$(\Phi, \mathcal{K})$, where~$\Phi$ is the base LLM and $\mathcal{K}$ is the library. Starting with an empty library~$\mathcal{K}_0=\emptyset$, the agent is given a sequence of tasks~$\{x_1, x_2, ..., x_T\}$, each sampled from the task pool~$\mathcal{T}$, and the library~$\mathcal{K}_t$ evolves with the history. At step~$t$, given the input problem~$x_t$ and the current library~$\mathcal{K}_{t-1}$, the agent samples abstractions from the library and generates a solution~$\tilde{y}_t$ for~$x_t$:
\begin{align*}
    &\{z_{t-1}\} \sim \mathrm{Sample}(\mathcal{K}_{t-1}, x_t), \\
    &\tilde{y}_t \sim P_\Phi(\cdot \mid x_t, \{z_{t-1}\}),
\end{align*}
then extracts new abstractions~$\{z_t\}$ from~$\tilde{y}_t$:
\begin{align*}
    \{z_t\} = \mathrm{Extract}(x_t, \tilde{y}_t).
\end{align*}
Finally, the agent updates the library based on the current library and this attempt:
$$
    \mathcal{K}_t = \mathrm{Update\_Library}[\mathcal{K}_{t-1}, (x_t, \tilde{y}_t, \{z_{t-1}\}, \{z_t\})].
$$

\subsection{Library Representation}
\label{sec:library}

The library $\mathcal{K}$ is represented as a weighted collection of knowledge abstractions. It includes two types of knowledge:~1) \emph{modular skills} that can be adapted to solve new tasks. For coding tasks, these are functions; for reasoning tasks, they are sub-problems with corresponding solutions extracted from complete solution trajectories; for agentic tasks, they are sub-tasks with corresponding workflows extracted from complete action trajectories. And~2) \emph{reflective insights} in natural language capturing lessons learned from past trajectories, including common mistakes and corrective strategies (e.g. ``If the task involves logistic regression, then do not forget to scale the features'').

\paragraph{Sampling.}
When sampling from the library to solve a new task, we first filter the library entries based on their embedding-based similarity to the task description and then use weighted sampling among the filtered entries based on their weights~$w(z \mid \mathcal{K})$. These weights are updated dynamically after generating new solutions for a problem based on the Information Gain and Future Information Gain described in \cref{sec:scoring}.

\subsubsection{Extracting Abstractions}
\label{sec:abstraction_extraction}
We extract \textit{modular skills}~$\{z_t\}$ directly from model outputs~$\tilde{y}_t$, as these skills are typically \textit{inherent} in the model's reasoning steps either explicitly or implicitly. For coding tasks, we extract functions as skills directly from the synthesized programs. For reasoning and agentic tasks, we extract skills by breaking down the reasoning steps or an action-observation sequence into several sub-modules and summarizing each sub-module into a new, self-contained skill through an LLM call.

To extract \textit{reflective insights}~$\{z_t\}$, we first use the LLM itself to estimate the score~$S_\Phi(\tilde{y}_t \mid x_t) \in [0, 1]$ for the model solution~$\tilde{y}_t$~(with optional textual feedback)~\cite{zheng2023judging,weng2023selfverification}. For coding tasks,~$S_\Phi(\tilde{y}_t \mid x_t)$ is obtained by instructing the LLM to generate synthetic test cases for the problem~$x_t$ and measuring the pass rate of the solution program~$\tilde{y}_t$ against the synthetic test. For reasoning tasks,~$S_\Phi(\tilde{y}_t \mid x_t)$ is obtained by majority voting~\cite{wang2023selfconsistency} and LLM-as-a-Judge~\cite{zheng2023judging} to break tie. For multi-turn agentic tasks, the score is obtained by passing the action-observation sequence~$\tilde{y}_t$ to the LLM and asking it to judge how many sub-goals in the task~$x_t$ have been accomplished. Next, we extract the insights by instructing the LLM to reflect on the solution~$\tilde{y}_t$ based on~$x_t$ and~$S_\Phi(\tilde{y}_t \mid x_t)$ and summarize it into several self-contained insights.

\begin{algorithm}[t]
\begin{algorithmic}[1]
\STATE \textbf{Input:} Task pool $\mathcal{T}$ of size~$N$, initial library $\mathcal{K}_0 = \emptyset$, model $\Phi$, number of iterations~$T$
\STATE Initialize best solutions $\{y^*_j\}_{j=1}^{N}$ for each task $x_j \in \mathcal{T}$
\FOR{$t$ from 1 to $T$}
    \STATE Draw a task $x_t \sim \mathcal{T}$ \COMMENT{Solution Generation}
    \FOR{$k$ from 1 to $K$}
        \STATE Sample $M$ abstractions $\{z_{t-1}^{(k)}\} \sim \mathrm{Sample}(\mathcal{K}_{t-1}, x_t)$
        \STATE Sample solution trajectory $\tilde{y}_t^{(k)} \sim P_\Phi(\cdot \mid x_t, \{z_{t-1}^{(k)}\})$
        \STATE Estimate the accuracy score $S_\Phi(\tilde{y}_t^{(k)} \mid x_t)$
        \IF{$S_\Phi(\tilde{y}_t^{(k)} \mid x_t) > S_\Phi(y^*_t \mid x_t)$}
            \STATE $y^*_t \leftarrow \tilde{y}_t^{(k)}$
        \ENDIF
    \ENDFOR
    \STATE $k^* \leftarrow \arg\max_k S_\Phi(\tilde{y}_t^{(k)} \mid x_t)$
    \STATE $\tilde{y}_t \leftarrow \tilde{y}_t^{(k^*)}$
    
    \STATE 
    \STATE $\mathcal{K}_{\mathrm{new}} \leftarrow \mathcal{K}_{t-1}$ \COMMENT{Library Update}
    \STATE Extract new abstractions $\{z_t\} \leftarrow \mathrm{Extract}(x_t, \tilde{y}_t)$
    \FOR{$z_t$ in $\{z_t\}$}
        \STATE Compute $\mathrm{IG}(z_t \mid x_t, \mathcal{K}_{t-1})$ based on ~\cref{eq:ig} if $z_t$ is a skill else 0
        \STATE $\mathcal{K}_{\mathrm{new}}\leftarrow \mathrm{Consolidate}[(z_t, \mathrm{IG}(z_t \mid x_t, \mathcal{K}_{t-1})), \mathcal{K}_{\mathrm{new}}]$
    \ENDFOR
    \FOR{$z_{t-1}$ in $\{z_{t-1}^{(k)}\}$}
        \STATE Compute $\mathrm{FutureIG}(z_{t-1} \mid x_t, \mathcal{K}_{t-1})$ based on ~\cref{eq:future_ig}
        \STATE $\mathcal{K}_{\mathrm{new}}\leftarrow \mathrm{Update\_FutureIG}[(z_{t-1}, \mathrm{FutureIG}(z_{t-1} \mid x_t, \mathcal{K}_{t-1})), \mathcal{K}_{\mathrm{new}}]$
    \ENDFOR
    \STATE $\mathcal{K}_t \leftarrow \mathcal{K}_{\mathrm{new}}$
\ENDFOR
\STATE \textbf{Output:} Final library $\mathcal{K}_T$, best solutions $\{y^*_j\}_{j=1}^{N}$
\end{algorithmic}
\caption{\codename}
\label{algorithm}
\end{algorithm}

\subsubsection{Weighting Mechanism}
\label{sec:scoring}
The core mechanism that enables evolution is credit propagation along the evolution chains of abstractions: useful abstractions not only receive direct reward~(through \textit{Information Gain}), but also propagate credit backward to the abstractions that enabled their creation~(through \textit{Future Information Gain}). Specifically, we focus on the evolution chains:
$$
\{ z_1 \} \xrightarrow{x_2, \tilde{y}_{2}} \{ z_2 \} \xrightarrow{x_3, \tilde{y}_{3}} ... \xrightarrow{x_{t-1}, \tilde{y}_{t-1}} \{ z_{t-1} \} \xrightarrow{x_t, \tilde{y}_{t}} \{ z_{t} \} \xrightarrow{x_{t+1}, \tilde{y}_{t+1}} ...,
$$
where each step of the evolution~$\{ z_{t-1} \} \xrightarrow{x_t, \tilde{y}_{t}} \{ z_{t} \}$ is realized by:
$$
\tilde{y}_{t} \sim P_\Phi(\cdot \mid x_{t}, \{ z_{t-1} \}),\, \{z_t\} = \mathrm{Extract}(x_t, \tilde{y}_t).
$$

If~$\{ z_{t} \}$ is proven to be useful for producing a high-quality solution for the task~$x_t$, the credit should be propagated not only to the new abstractions~$\{ z_{t} \}$, but also to the preceding abstractions~$\{ z_{t-1} \}, \{ z_{t-2} \}, ..., \{ z_{1} \}$ along the evolution chain. In practice, to balance between efficacy and computational cost, we propagate the credit by two steps, i.e. to~$\{ z_{t} \}$ and~$\{ z_{t-1} \}$.

More formally, to propagate the credit to~$\{ z_{t} \}$, we first define the baseline score for~$x_t$ given~$\mathcal{K}_{t-1}$:
\begin{align}
    \mu_{\mathrm{base}}(Y \mid x_t, \mathcal{K}_{t-1})
    = \mathbb{E}_{\tilde{y}_{t}, Z_{t-1}}
    \left[S_\Phi(\tilde{y}_{t} \mid x_t)\right],
\end{align}
where~$\tilde{y}_{t}$ and~$Z_{t-1}$ follow~$\tilde{y}_{t} \sim P_\Phi(\cdot \mid x_{t}, Z_{t-1})$ and~$Z_{t-1}\sim\mathrm{Sample}(\mathcal{K}_{t-1}, x_t)$ in the expectation.

We then define the conditional score given that~$z_t$ is present in the extracted abstractions:%
\begin{align}
    \mu_{\mathrm{cond}}(Y \mid x_t, \mathcal{K}_{t-1}, z_t)
    = \mathbb{E}_{\tilde{y}_{t}, Z_{t-1}}
    \left[S_\Phi(\tilde{y}_{t} \mid x_t) \mid z_t \in \mathrm{Extract}(x_t, \tilde{y}_t) \right].
\end{align}
Next, we define \textbf{Information Gain~(IG)}, which measures how much better the model~$\Phi$ can solve~$x_t$ when~$z_t$ is present in the extracted abstractions compared to the average baseline:
\begin{equation}
    \mathrm{IG}(z_t \mid x_t, \mathcal{K}_{t-1})
    = \log\big(\mu_{\mathrm{cond}}(Y \mid x_t, \mathcal{K}_{t-1}, z_t)\big)
    - \log\big(\mu_{\mathrm{base}}(Y \mid x_t, \mathcal{K}_{t-1})\big).
\label{eq:ig}
\end{equation}
In practice, at each iteration we sample~$K$ solutions for~$x_t$, each conditioned on abstractions sampled from the library~(\cref{algorithm}). We estimate~$\mu_{\mathrm{base}}$ by the average score over all~$K$ samples, and~$\mu_{\mathrm{cond}}$ by the average score over the samples from which the exact abstraction~$z_t$ is extracted.

Furthermore, to propagate the credit back to~$\{ z_{t-1} \}$, we first define the baseline score where~$z_{t-1}$ is not present in the previously sampled abstractions:\footnote{Note that IG uses the unconditional baseline, while Future IG uses an exclusion baseline to avoid bias introduced by repeated sampling of the same abstraction.}
\begin{align}
    \mu_{\mathrm{base}}(Y \mid x_t, \mathcal{K}_{t-1}, \overline{z_{t-1}})
    = \mathbb{E}_{\tilde{y}_{t}, Z_{t-1}}
    \left[S_\Phi(\tilde{y}_{t} \mid x_t) \mid z_{t-1} \notin Z_{t-1}\right],
\end{align}
and the conditional score where~$z_{t-1}$ is present:
\begin{align}
    \mu_{\mathrm{cond}}(Y \mid x_t, \mathcal{K}_{t-1}, z_{t-1})
     = \mathbb{E}_{\tilde{y}_{t}, Z_{t-1}}
    \left[S_\Phi(\tilde{y}_{t} \mid x_t) \mid z_{t-1} \in Z_{t-1}\right].
\end{align}

And finally, we define \textbf{Future IG}, which quantifies how much better the model can solve~$x_t$ given~$z_{t-1}$ as one of the preceding abstractions, compared to the baseline score:
\begin{equation}
    \mathrm{FutureIG}(z_{t-1} \mid x_t, \mathcal{K}_{t-1})
    = \log\big(\mu_{\mathrm{cond}}(Y \mid x_t, \mathcal{K}_{t-1}, z_{t-1})\big)
    - \log\big(\mu_{\mathrm{base}}(Y \mid x_t, \mathcal{K}_{t-1}, \overline{z_{t-1}})\big).
\label{eq:future_ig}
\end{equation}
 
Future IG can be interpreted as estimating the contribution of an abstraction to the generation of useful future abstractions.

Overall, we compute the weight of each abstraction~$z$ in the library based on its IG and Future IG scores:
\begin{equation}
    w(z \mid \mathcal{K}) = \tau \max_{x \in \mathcal{T}} \mathrm{IG}(z \mid x, \mathcal{K}) + \mathbb{E}_{x \sim \mathcal{T}} \mathrm{FutureIG}(z \mid x, \mathcal{K}),
\label{eq:weight}
\end{equation}
where $\tau$ is a hyperparameter controlling the relative importance of immediate versus future gain. 
In practice, we use~$\tau=1$ for all skills and~$\tau=0$ for all insights, as skills can directly contribute to solution construction, whereas insights may not. Future IG is computed for all abstractions (including skills and insights), as both can influence future solution and abstraction generation.

We use the maximum IG over tasks to capture the peak utility of an abstraction, encouraging any abstraction with immediate utility to be sampled and evolve in the future. For Future IG, we take the expectation of Future IG over tasks randomly sampled from the task pool by maintaining a list of Future IG scores for each~$z$, initialized by an empty list. At each iteration, when~$z_{t-1}$ is sampled from the library and integrated in the context to solve a task~$x_t$, we compute~$z_{t-1}$'s Future IG given~$x_t$ based on~\cref{eq:future_ig} and add it to its Future IG list. When computing~$w(z \mid \mathcal{K})$, we estimate the expected Future IG by averaging over the Future IG scores in the list.

Updating Future IG dynamically in this manner makes the agent less prone to the noise in self-evaluated scores, as we take the average over Future IG scores across iterations. It also enables continual evolution of the library \---\ when a more advanced abstraction covering a broader range of use cases emerges, it typically receives high Future IG, while older, less general abstractions exceeded by the new abstraction will receive decreasing Future IG. As a result, more advanced abstractions gradually become more prevalent with higher sampling probability, while older ones are progressively de-emphasized without explicit removal.

\subsection{Library Update}
\label{sec:library_update}

Building on the abstraction extraction~(\cref{sec:abstraction_extraction}) and weighting mechanism~(\cref{sec:scoring}), we now describe how the library is updated.
For each abstraction in the library, we maintain its IG score and a list of Future IG scores obtained on previous tasks.
The library is updated in two ways:~1) \textbf{abstraction consolidation} where new abstractions are consolidated with similar entries and added to the library, and~2) \textbf{updating Future IG} where for each abstraction integrated in the context for solving the current task, we update its Future IG list with the new Future IG score.

\paragraph{Abstraction consolidation.}
When a new abstraction is extracted from a trajectory, we first retrieve the most similar abstraction of the same type from the library using embedding-based similarity. We then prompt the LLM to consolidate the new abstraction into the most similar one from the library when the abstractions are semantically and functionally similar. If the two abstractions can be merged, the IG score of the consolidated entry is set to the average of the IG scores of the new and the existing abstractions, and the Future IG list of the existing entry is carried over to the consolidated entry. Otherwise, the new abstraction is added as a separate entry~(with the current IG score and an empty Future IG list) into the library. This consolidation step is critical for promoting the emergence of abstractions that generalize beyond specific problems.

\cref{algorithm} summarizes the full procedure, alternating between solution generation and library update~(as illustrated in \cref{fig:algorithm}).
In the solution generation step, the agent samples a set of abstractions from the current library to construct a contextual prompt, generates and estimates the scores of candidate solutions. In the library update step, newly extracted abstractions are consolidated into the library, and the Future IG scores of the sampled abstractions are updated. The agent iterates through the two steps across successive tasks, enabling the library to continuously grow and evolve.

\section{Experiments}
\label{sec:experiments}

\subsection{Experimental Setup}
\paragraph{Evaluation datasets.}
We evaluate \codename on a diverse set of benchmarks, including mathematical reasoning, code generation, and multi-turn agentic tasks, in two different settings:~1) static benchmarks, where the agent learns through multiple iterations over the whole dataset, and~2) continual learning, where tasks are presented in a stream and each task is attempted only once. In both settings, the agent learns in a self-supervised manner without access to the ground-truth answers or tests.

For the static setting, we evaluate it on challenging math and coding benchmarks including~a) exam questions from Harvard-MIT Mathematics Tournament~(\textbf{HMMT}) 2025--2026, which is an elite high-school competition featuring complex mathematical reasoning, b) \textbf{BigCodeBench Hard}~\citep{zhuo2024bigcodebench}, which contains 148 challenging programming tasks designed to evaluate compositional reasoning and the use of existing function calls for real-world programming, and~c) \textbf{LiveCodeBench v6 Hard}~\citep{jain2024livecodebench}, which contains 80 highly challenging competitive coding problems.

For the continual learning setting, we evaluate it on multi-turn agentic tasks, including the \textbf{ScienceWorld}~\cite{wang-etal-2022-scienceworld} and \textbf{PDDL}~\cite{vallati2014pddl} tasks from the AgentBoard benchmark suite~\citep{ma2024agentboard}. These tasks require multi-round interaction with a partially observable environment. We adopt the task order in the original dataset in our evaluation. See more dataset details in \cref{sec:dataset_details}.

\paragraph{LLMs.}
We evaluate the efficacy of \codename using both general-purpose LLMs, such as GPT-4o (on BigCodeBench), and LLMs featuring effective reasoning and coding, such as o4-mini (on HMMT, LiveCodeBench, ScienceWorld, and PDDL). On tasks that require intensive reasoning, such as HMMT and LiveCodeBench, we use o4-mini with high reasoning effort, while on multi-turn agentic tasks, we use o4-mini with low reasoning effort. We additionally evaluate \codename with the open-weight Qwen3.5-9B~\citep{qwen3.5} on BigCodeBench.

\paragraph{Baselines.}

We compare \codename against both test-time scaling~(TTS) and test-time learning~(TTL) approaches. For TTS, we adopt
\begin{inparaenum}[1)]
    \item \textbf{Best-of-N}, which generates~$N$ independent solutions and selects the one with the highest estimated accuracy, and
    \item \textbf{Recursive Self-Aggregation~(RSA)}~\citep{venkatraman2025rsa}, which iteratively refines and improves a population of candidate solutions through aggregation of subsets.\footnote{We exclude the TTS baselines on the agentic tasks, as each task is attempted only once through multiple rounds.}
\end{inparaenum}
For TTL, we include
\begin{inparaenum}[1)]
    \item \textbf{ExpRAG}~\citep{wei2025evomemory}, an agent augmented with a memory of past trajectories,
    \item \textbf{ReasoningBank}~\citep{ouyang2026reasoningbank}, which distills reasoning strategies from self-judged successful and failed trajectories into a growing memory, and
    \item \textbf{Dynamic Cheatsheet~(DC)}~\citep{suzgun2025dc}, which augments an LLM with an evolving textual memory of procedural knowledge.\footnote{We choose ReasoningBank over AWM~\citep{wang2025awm}, because it is reported to consistently outperform AWM~\citep{ouyang2026reasoningbank}. As ReasoningBank is designed for multi-turn agentic tasks, we evaluate it on the AgentBoard tasks.}
\end{inparaenum}
TTS methods solve each task independently without access to experience from other tasks, whereas the TTL baselines exploit prior experience from the same test stream and thus serve as controlled comparisons for history access.

\paragraph{Evaluation metrics.}
On HMMT, we evaluate the accuracy of the LLM's solutions against the ground-truth answers using the grading script in~\cite{balunovic2025matharena}. On BigCodeBench and LiveCodeBench, we measure task performance by the pass rate on the true test cases for each problem. On ScienceWorld and PDDL, we adopt the success rate and progress rate from the AgentBoard~\citep{ma2024agentboard}. On each benchmark, we compare the performance of various methods under the same cost budget measured by weighted token count, where the input token count is weighted by~$\lambda=1$ while the output token count is weighted by~$\lambda=4$.\footnote{The weights are based on the typical per-token rates of input and output tokens for proprietary LLMs.}

\subsection{Main Result: \codename Consistently Outperforms Baselines}

\begin{table}[t]
\centering
\caption{Performance comparing \codename against base sampling, test-time scaling~(TTS) methods, including Best-of-N and Recursive Self-Aggregation~(\textit{RSA}), along with test-time learning~(TTL) baselines, including ExpRAG, ReasoningBank~(\textit{RB}), and Dynamic Cheatsheet~(\textit{DC}), across math~(\textit{HMMT}), coding~(\textit{BigCodeBench} and \textit{LiveCodeBench}), and multi-turn agentic tasks~(\textit{ScienceWorld} and \textit{PDDL}). For each task, we report the average score over three runs for each method. A dash~(--) indicates the method is not applicable to that setting. An asterisk~(*) denotes that the best score(s) is significantly higher than the second-best on each benchmark based on paired student's t-test with~$p<0.05$.}
\label{tab:main}
\small
\scalebox{0.95}{
\begin{tabular}{@{}lccccccc@{}}
\toprule
\multirow{2}{*}{\textbf{Method}} & \textbf{HMMT} & \textbf{BigCodeBench} & \textbf{LiveCodeBench} & \multicolumn{2}{c}{\textbf{ScienceWorld}} & \multicolumn{2}{c}{\textbf{PDDL}} \\
\cmidrule(lr){5-6} \cmidrule(lr){7-8}
 & Accuracy & Pass Rate & Pass Rate & Success & Progress & Success & Progress \\
\midrule
Base & 57.0 & 29.7 & 58.8 & 53.3 & 82.9 & 62.8 & 79.1 \\
\midrule
\multicolumn{8}{@{}l}{\textit{Test-Time Scaling (TTS)}} \\
Best-of-N & 74.2 & 37.2 & 67.5 & -- & -- & -- & -- \\
RSA~\citep{venkatraman2025rsa} & 67.2 & 31.7 & \textbf{70.0*} & -- & -- & -- & -- \\
\midrule
\multicolumn{8}{@{}l}{\textit{Test-Time Learning (TTL)}} \\
ExpRAG~\citep{wei2025evomemory} & 60.2 & 25.0 & 51.3 & 51.1 & 80.4 & 39.4 & 56.7 \\
RB~\citep{ouyang2026reasoningbank} & -- & -- & -- & 56.7 & 82.8 & 62.8 & 78.5 \\
DC~\citep{suzgun2025dc} & 66.7 & 35.8 & 65.0 & 55.9 & 83.6 & 65.0 & 83.6 \\
\midrule
\codename & \textbf{77.4*} & \textbf{40.8*} & \textbf{70.0*} & \textbf{57.4} & \textbf{84.3} & \textbf{72.8*} & \textbf{86.2} \\
\bottomrule
\end{tabular}
}
\end{table}

\begin{figure}[t]
    \begin{subfigure}[b]{0.49\textwidth}
        \includegraphics[width=\textwidth]{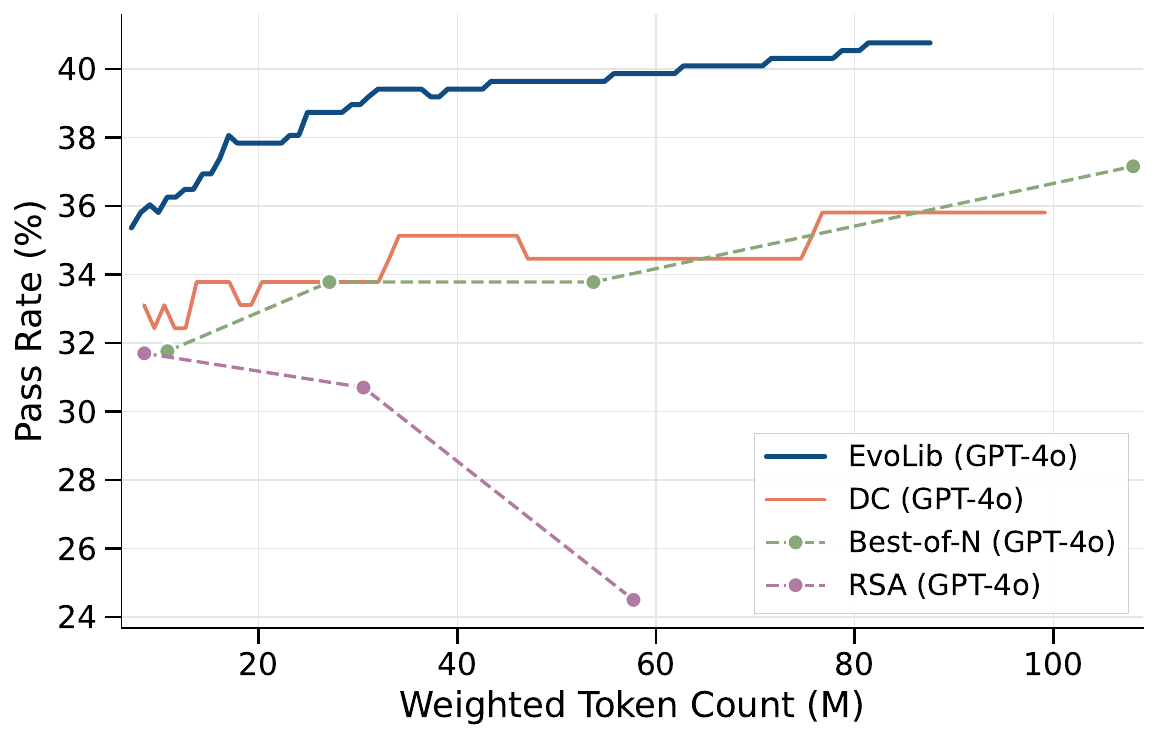}
        \caption{BigCodeBench}
    \end{subfigure}
    \begin{subfigure}[b]{0.49\textwidth}
        \includegraphics[width=\textwidth]{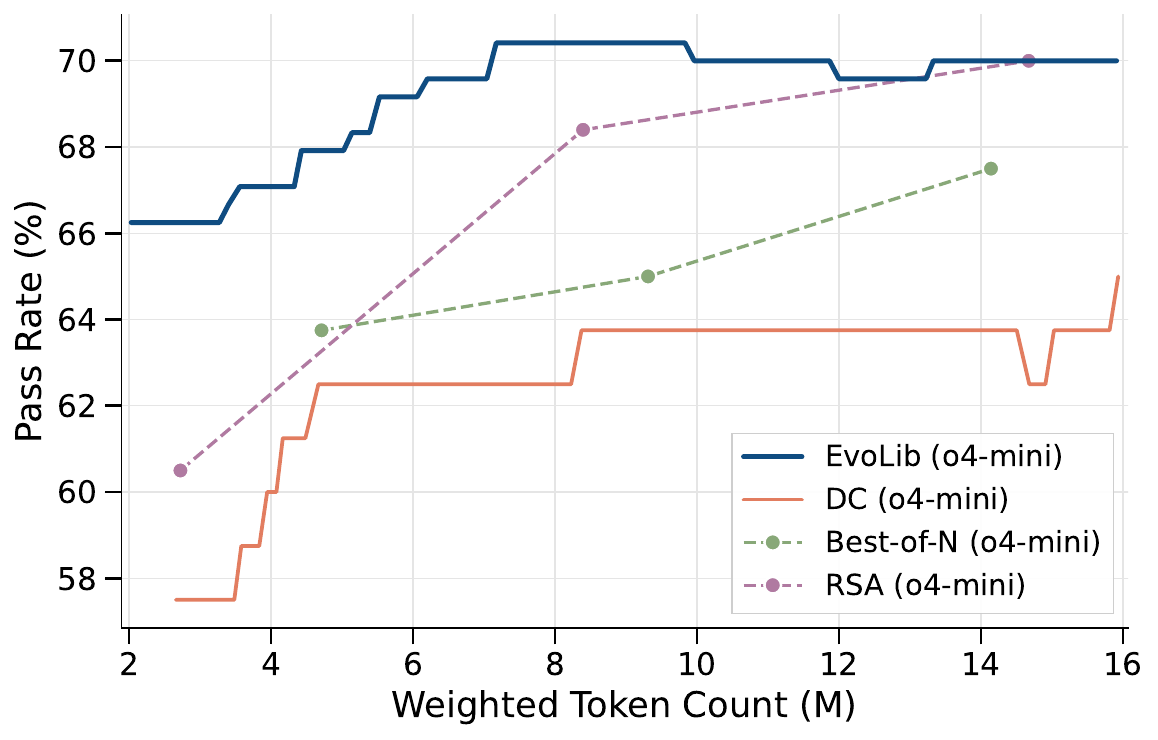}
        \caption{LiveCodeBench}
    \end{subfigure}
\caption{Cost–performance curves comparing \codename with competitive baselines on (a) BigCodeBench and (b) LiveCodeBench. Each curve plots performance~(y-axis) as the test-time compute cost~(x-axis) increases for each method.}
\label{fig:main_cost_performance}
\end{figure}

\paragraph{Comparison on static benchmarks.}
\cref{tab:main} shows that \codename achieves the strongest average performance across all evaluated settings, including math, code generation, and multi-turn agentic tasks.
In the static setting, \codename improves over base sampling by~11-20\%, while achieving better or competitive performance compared with the best TTS baseline. On BigCodeBench, GPT-4o augmented with \codename even surpasses the strongest reported LLMs~(e.g., \emph{Gemini-Exp-1206} at 40.5\%) on the leaderboard. This suggests that test-time learning can substantially close, and in some cases overcome, the gap between less and more powerful base models.

Furthermore, as shown in the cost-performance curves in~\cref{fig:main_cost_performance,fig:main_cost_performance_hmmt}, \codename is more token-efficient than both TTS baselines \---\ it yields larger performance gains under tighter cost budgets~(+6-16\% on BigCodeBench and~+3-6\% on LiveCodeBench when the budget is reduced by half).
These gains indicate that learning and sharing knowledge across problem instances yields consistent improvements over TTS.

Compared to TTL baselines, \codename consistently outperforms DC, the best TTL baseline on static benchmarks, by~5-11\% across three benchmarks, and is also more token-efficient~(\cref{fig:main_cost_performance}). 

\paragraph{Comparison in continual learning setting.}
\cref{tab:main} also shows that \codename outperforms all existing TTL methods in the continual learning setting, including ExpRAG, ReasoningBank, and DC, with the largest gain on PDDL~(+10\% and +8\% success rate over ReasoningBank and DC). These results demonstrate that \codename transfers what it has learned on past tasks to future tasks more effectively than the existing TTL methods.

\paragraph{Cost breakdown.}
Solution generation accounts for most of \codename's LLM calls~(54--87\%), with the remainder spent on scoring, abstraction extraction, and consolidation. See \cref{sec:cost_details} for raw token counts and LLM call breakdown.

\begin{figure}[t]
    \begin{subfigure}[b]{0.49\textwidth}
        \includegraphics[width=\textwidth]{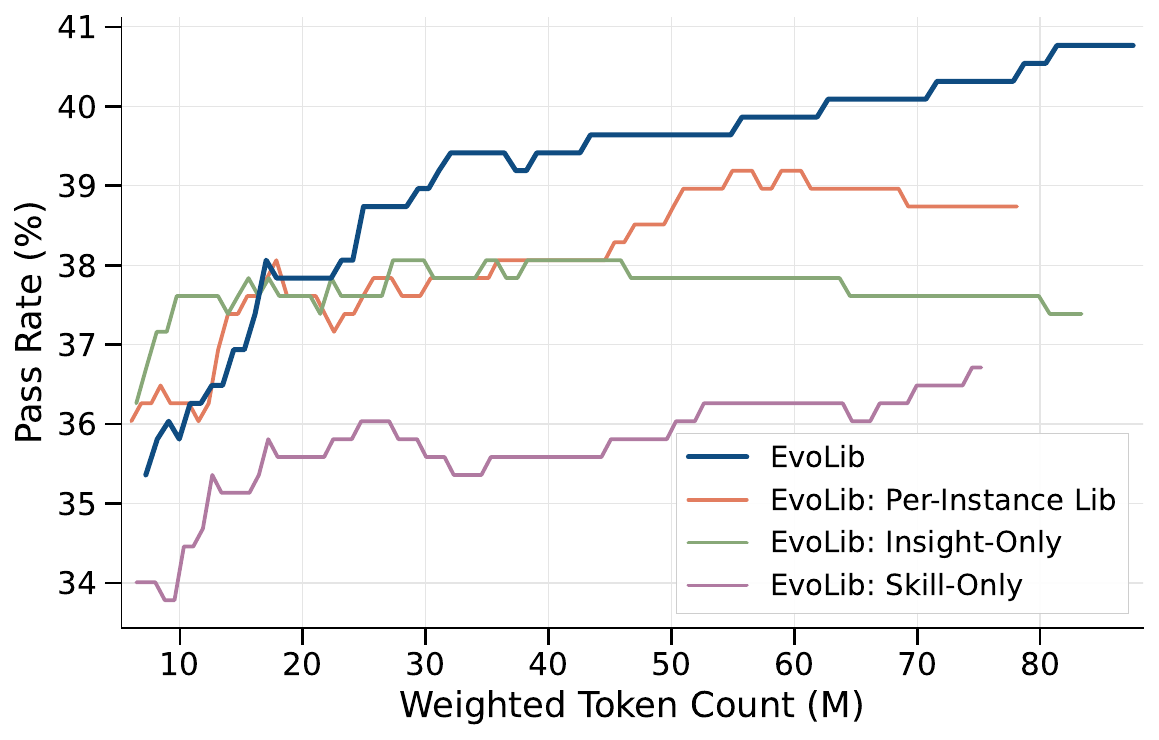}
        \caption{Ablating Abstraction Types}
    \end{subfigure}
    \begin{subfigure}[b]{0.49\textwidth}
        \includegraphics[width=\textwidth]{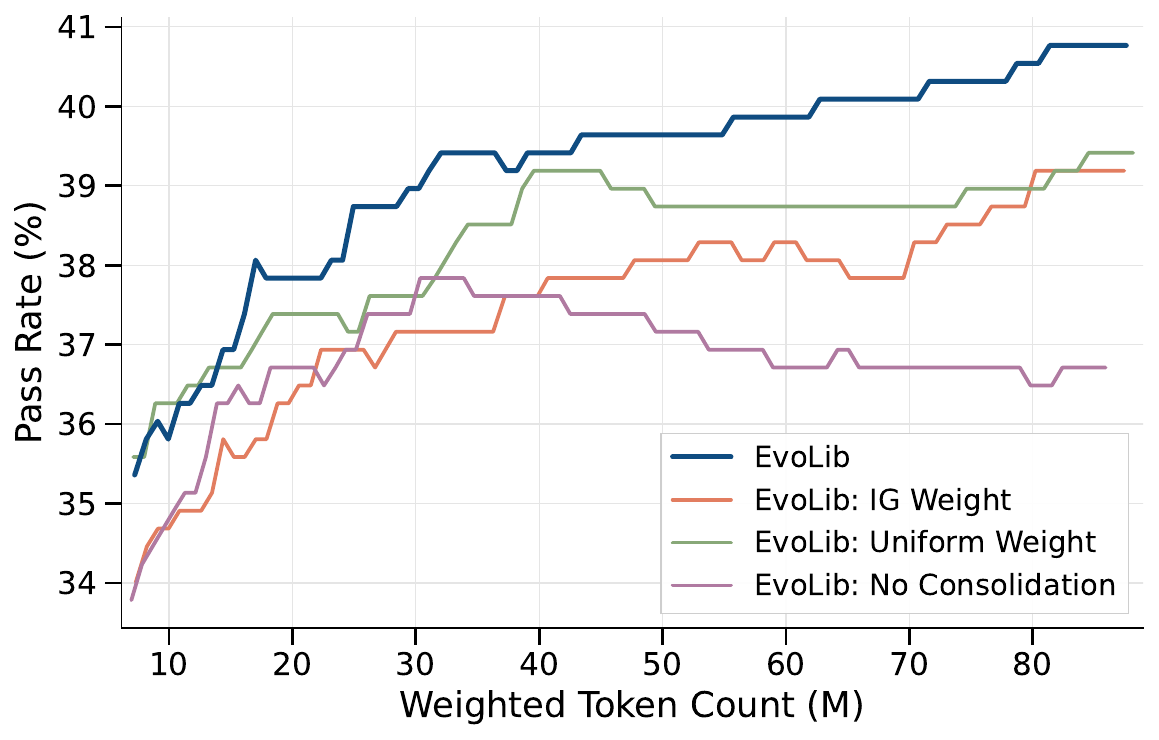}
        \caption{Ablating Weighting and Consolidation}
    \end{subfigure}
\caption{Ablation study on BigCodeBench. (a) Comparison of variants using different abstraction types or per-instance libraries. (b) Comparison of variants with different weighting strategies and without consolidation. Each plot shows the performance in pass rate~(y-axis) as a function of test-time compute cost~(x-axis).}
\label{fig:ablation}
\end{figure}

\subsection{Ablation Study}
\label{sec:ablation}

\paragraph{Both abstraction types and cross-instance sharing matter.}
\cref{fig:ablation}(a) ablates the types of abstractions stored in the library and how the library is shared.
Using only modular skills or only reflective insights underperforms the full variant, which suggests that the two abstraction types contribute complementary value: insights capture higher-level corrective signals that guide future attempts~(especially useful at the early stage), while skills provide solution components that can be reused and recombined to solve more complex tasks.
In addition, the per-instance variant lags behind the shared library, especially at larger budgets, which provides strong evidence that the library generalizes beyond problem-specific details. It also supports the claim that cross-problem knowledge reuse and evolution, rather than iterative knowledge refinement within each problem, drives further performance gain.

\paragraph{Abstraction consolidation leads to substantial gain.}
\cref{fig:ablation}(b) shows the impact of abstraction consolidation. Disabling consolidation hurts performance, particularly as compute increases. We further analyze the role of abstraction consolidation in \cref{sec:extended_results}. \cref{fig:lib_size_curve,tab:consolidation_cases} indicate that the consolidation step helps control memory growth and turns task-specific abstractions into more generic ones that apply across multiple problems.

\paragraph{Weighting and consolidation are both needed for sustained improvement.}
\cref{fig:ablation}(b) also shows the impact of IG and Future IG scores in the weighting mechanism.
Uniform weighting leads to weaker scaling at higher compute budgets, where the library grows larger and effective sampling becomes increasingly important. Removing the Future IG score also degrades performance substantially, as it is crucial for selecting abstractions that can evolve into more advanced ones.

\subsection{Robustness Analysis}
\label{sec:robustness}

\paragraph{Robustness to task ordering.}
\label{sec:task_ordering}
We study robustness to task ordering in the continual learning settings on PDDL, which contains multiple distinct task types~(see \cref{tab:pddl_order} and discussion in \cref{paragraph:task_order} for details). We find that \codename consistently outperforms DC across easy-to-hard, hard-to-easy, and random task streams, with the largest gain~(+9\%) under random ordering. The result indicates that \codename can continually learn from diverse tasks even when they are interleaved, suggesting its practical advantage in real-world scenarios where an agent must handle and learn from a mixed stream of heterogeneous user requests without relying on a structured curriculum.

\paragraph{Robustness to self-evaluation noise.}
\label{sec:self_eval}
\codename does not have access to the ground-truth feedback at test time. Instead, it updates the library based on a self-evaluation signal, such as whether a solution passes model-generated tests. This signal can be noisy, and its noise level varies widely across benchmarks~(\cref{sec:self_eval_details}). It is nearly perfect on PDDL~(ROC-AUC 99.3), but much noisier on the coding benchmarks~(ROC-AUC 67.3 on BigCodeBench and 79.5 on LiveCodeBench). Even so, \codename remains effective where the signal is noisy. On BigCodeBench and LiveCodeBench, it still improves over base sampling by about~+11 points and matches or outperforms all baselines~(\cref{tab:main}). The gain also holds up with a weaker judge: when Qwen3.5-9B is used to generate the synthetic tests on BigCodeBench, \codename still outperforms Best-of-N under the same judge~(35.4\% vs.\ 33.8\%). We attribute this robustness to the weighting mechanism. Each abstraction's Future IG score is averaged over many iterations rather than determined by any single self-evaluated outcome, so occasional misjudgments have limited effect on which abstractions are retained and reused.

\paragraph{Open-weight model results.}
\label{sec:open_weight}
With the weaker, open-weight Qwen3.5-9B as the base model on BigCodeBench, \codename achieves a pass rate of 33.1\%, outperforming Best-of-N~(27.0\%) by~6.1 points, which suggests that its benefits transfer to weaker open-weight models.

\section{Discussion}
\label{sec:discussion}

This paper introduces \codename, a framework that enables black-box LLMs to self-improve at test time through abstraction induction, reuse, and evolution.
A key property of \codename is its favorable scaling behavior: performance improves consistently as more problems are solved and more compute is invested, without external supervision. This draws a direct analogy to human cognitive development. Just as humans build progressively more abstract mental models through experience, the library evolves from concrete, task-specific functions toward more general, reusable abstractions. 
Experiments on mathematical reasoning, code generation, and interactive agentic tasks show that, under matched compute budgets, \codename outperforms strong test-time scaling and test-time learning methods, with gains that persist under noisy self-evaluation.
The framework offers a promising direction for LLMs to learn and self-improve during deployment without requiring gradient-based fine-tuning or supervision signals.

\paragraph{Limitations.}
\label{sec:limitations}

The current \codename framework assumes that the model's own judgment of a solution provides more useful signals than noise for self-evaluation. Thus, the method works best for tasks in which verifying a solution is easier than generating one (e.g., coding tasks with executable test cases). For tasks where verification requires comparable or even greater effort than solution generation, a powerful external verifier model may be necessary to reliably assess each candidate solution.
Additionally, the current evaluation focuses on math, coding, and agentic tasks. It is unclear how well the approach generalizes to other domains, such as open-ended question-answering or scientific tasks, with different knowledge types and feedback characteristics.

\paragraph{Future work.}
\label{sec:future_work}
One promising direction is to enable LLMs to meta-learn the design of the library itself, including the choice of abstraction forms, the weighting mechanism, and the library update algorithm. Rather than fixing these components, the system could optimize them over time based on previous tasks, sampled trajectories, and the structure of the task stream, potentially leading to more efficient test-time learning schemes.
Another direction is to extend \codename to open-ended domains, such as open math problems or challenging optimization tasks, by incorporating sparse external evaluation or human feedback to drive a self-improving verifier. 
Since the library is stored explicitly, it can also be inspected, edited, or reset by humans. Studying such human oversight of the evolving library is an interesting avenue for future work.

\small
\bibliography{reference.bib}
\bibliographystyle{unsrt}

\appendix
\newpage
\counterwithin{figure}{section}
\counterwithin{table}{section}

\section{Experimental Setup Details}
\label{sec:setup_details}

\begin{table}[t]
\centering
\caption{Summary of datasets used in evaluation.}
\small
\begin{tabular}{l l c l}
\toprule
\textbf{Domain} & \textbf{Dataset} & \textbf{\# Instances} & \textbf{Notes} \\
\midrule

\multirow{3}{*}{Math Reasoning}
& HMMT Feb 2025  & 30 & Competitive math problems \\
& HMMT Nov 2025  & 30 & Competitive math problems \\
& HMMT Feb 2026  & 33 & Competitive math problems \\
\cmidrule(lr){2-4}
& \textbf{HMMT (Total)} & 93 & Competitive math problems \\

\midrule

\multirow{2}{*}{Code Generation}
& BigCodeBench Hard & 148 & Real-world programming problems \\
& LiveCodeBench v6 (Hard) & 80 & Competitive programming problems \\

\midrule

\multirow{2}{*}{Agentic Tasks}
& ScienceWorld & 90 & Scientific experiment tasks \\
& PDDL & 60 & Symbolic planning tasks \\

\bottomrule
\end{tabular}
\label{tab:dataset_sizes}
\end{table}

\subsection{Dataset Details}
\label{sec:dataset_details}

We evaluate our method on a diverse set of benchmarks spanning mathematical reasoning, code generation, and multi-turn agentic tasks. All datasets used in this work are publicly available.

For mathematical reasoning, we use exam problems from the \texttt{Harvard--MIT Mathematics Tournament (HMMT)}, an elite high-school competition featuring challenging mathematical reasoning questions spanning algebra, combinatorics, geometry, and number theory. Specifically, we combine problem sets from three recent competitions: February 2025, November 2025 and February 2026, all obtained from the \texttt{MathArena} collection on HuggingFace.\footnote{\url{https://huggingface.co/datasets/MathArena/hmmt_feb_2025}, \url{https://huggingface.co/datasets/MathArena/hmmt_nov_2025}, \url{https://huggingface.co/datasets/MathArena/hmmt_feb_2026}}
We evaluate performance using accuracy against ground-truth answers, following the official grading scripts provided by \texttt{MathArena}.

For code generation, we evaluate on the \texttt{BigCodeBench Hard} benchmark, a subset of BigCodeBench designed to assess real-world programming and compositional reasoning abilities,\footnote{\url{https://huggingface.co/datasets/bigcode/bigcodebench-hard}} and \texttt{LiveCodeBench}, a benchmark of competitive programming problems.\footnote{\url{https://github.com/LiveCodeBench/LiveCodeBench} (we use the hard subset of v6).}
Performance is measured using pass@1-style accuracy (pass rate), i.e., whether the generated program passes all ground-truth~(including hidden) test cases.

For evaluating multi-turn agentic behavior, we use tasks from the AgentBoard benchmark suite.\footnote{\url{https://hkust-nlp.github.io/agentboard/}} Specifically, we evaluate on \texttt{ScienceWorld}, a simulated environment requiring sequential decision-making and scientific reasoning, and \texttt{PDDL}, symbolic planning tasks defined in the Planning Domain Definition Language.
Each task involves multi-step interaction with an environment, where the model generates actions conditioned on observations. Tasks are partially observable and require planning over multiple turns. We evaluate performance using the success rate (whether the task is completed) and progress rate (fraction of sub-goals achieved) provided by AgentBoard.

\cref{tab:dataset_sizes} summarizes the information of each dataset.

The static setting assumes that a task drawn from a benchmark shares similar solution strategies or knowledge with some other tasks in the same benchmark, so that a library learned on the benchmark can generalize to tasks from the same or a similar distribution. In the continual learning setting, each task is attempted only once in stream order, and the goal is to continuously adapt to new tasks at test time.

\subsection{Model Details}
We use GPT-4o~(version \texttt{2024-11-20})~\cite{hurst2024gpt4o} with a temperature of~0 and~$top\_p=0.5$ on BigCodeBench.
On ScienceWorld and PDDL, we use o4-mini~(version \texttt{2025-04-16}) with low reasoning effort.\footnote{\url{https://openai.com/index/o3-o4-mini-system-card/}}
On LiveCodeBench and HMMT, we use o4-mini~(version \texttt{2025-04-16}) with high reasoning effort.
In addition, we use the open-weight Qwen3.5-9B~\cite{qwen3.5} on BigCodeBench in two settings: 1)~as the base LLM for all operations~(\cref{sec:open_weight}), and 2)~as a weaker judge that generates the synthetic tests, while GPT-4o is used for solution generation and all library-related operations~(\cref{sec:self_eval_details}).

\subsection{Baseline Details}

We compare our method against representative test-time scaling (TTS) and test-time learning (TTL) methods. For TTS, we include Best-of-$N$ sampling, which generates $N$ independent candidate solutions per task and selects the one with the highest self-estimated score. We also evaluate Recursive Self-Aggregation (RSA)~\cite{venkatraman2025rsa}, a stronger TTS method that iteratively refines a population of reasoning trajectories by aggregating subsets of candidates, enabling progressive improvement across iterations. We run it with the population size~$N = 16$ and aggregation subset size~$K = 4$ with GPT-4o and o4-mini to balance between inference cost and performance. For TTL baselines, we include ExpRAG~\cite{wei2025evomemory},\footnote{We set the number of retrieved entries to~$4$ as recommended and use \emph{text-embedding-3-small} as the embedding model.} which augments the base model with a global memory of raw trajectories, and Dynamic Cheatsheet~\cite{suzgun2025dc},\footnote{We use the \emph{DC-Cumulative} variant in all our experiments as it outperforms the other variant in the preliminary experiment.} which maintains an adaptive memory~(i.e. a cheatsheet) of high-level principles and procedural knowledge extracted through self-reflection.
On the agentic tasks, we further include ReasoningBank~\cite{ouyang2026reasoningbank}, which distills generalizable reasoning strategies from both self-judged successful and failed trajectories into a memory and retrieves relevant entries for new tasks. We choose ReasoningBank over AWM~\cite{wang2025awm} because ReasoningBank is reported to consistently outperform AWM by learning from both successful and failed trajectories~\cite{ouyang2026reasoningbank}.
For all baselines, we run them with varying token cost budget~(i.e. varying~$N$ for Best-of-$N$ and varying number of iterations for the other methods) and compare performance under the same cost budget.
Best-of-$N$, ExpRAG, and DC use the same synthetic-test-based self-verification as \codename and are compared under the same budgets, so the controlled difference is the evolving library itself.
Note that Best-of-$N$ and RSA solve each task independently and are not history-matched; we include them to compare test-time learning against competitive TTS methods. The controlled comparisons for access to the test-stream history are ExpRAG, DC, and ReasoningBank, all of which exploit prior experience from the same test stream.

\subsection{\codename Details}
To estimate the baseline and conditional scores, the agent typically samples multiple times for each problem~$x_t$, except for the continual learning experiments where the agent attempts each task only once. In this case, the agent uses samples at previous turns for the same task to estimate the scores.
We set the number of trials~$K=3$ on code generation tasks, and~$K=10$ for mathematical reasoning, while on multi-turn agentic tasks, each task is attempted only once~(following the continual learning setting). To retrieve similar functions for abstraction consolidation, we set the embedding-based similarity threshold to~$0.8$. For weighted sampling, we sample at most~$10$ insights and~$10$ skills into the LLM prompt, and set the task-abstraction similarity threshold to~$0.2$ on agentic tasks and to~$0$ on the other benchmarks~(since the problems within each benchmark are highly related on these benchmarks). We use \textit{text-embedding-3-small} as the embedding model.

\paragraph{Hyperparameter sensitivity.}
We evaluate the sensitivity of \codename to its main hyperparameters on BigCodeBench. Performance varies by at most two points across different numbers of trials~($K \in \{3, 6, 10\}$) and different numbers of sampled skills and insights~($\{5, 10, 20\}$), while the consolidation similarity threshold~($\{0.7, 0.8, 0.9\}$) has a larger effect of approximately four points.

\section{Extended Results}
\label{sec:extended_results}
\counterwithin{figure}{section}
\counterwithin{table}{section}

\begin{figure}[t]
    \centering
    \includegraphics[width=.7\linewidth]{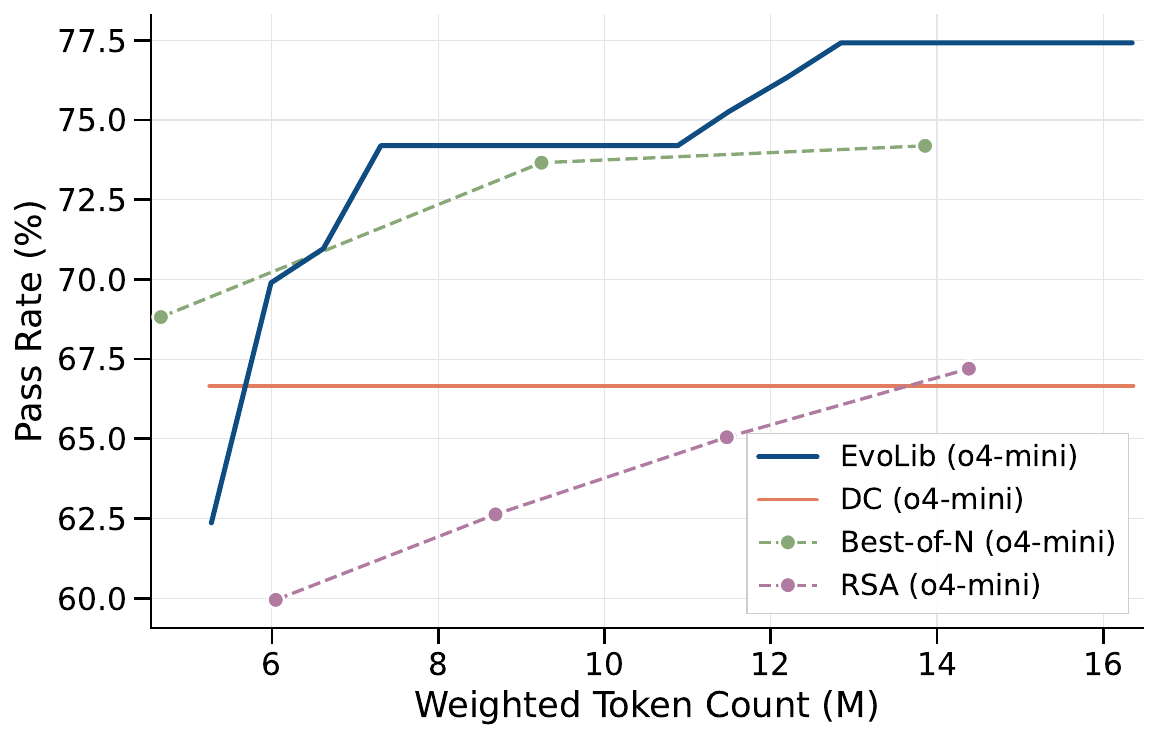}
\caption{Cost–performance curve comparing \codename with competitive baselines on HMMT. Each curve plots performance~(y-axis) as the test-time compute cost~(x-axis) increases for each method.}
\label{fig:main_cost_performance_hmmt}
\end{figure}

\begin{table}[t]
\centering
\caption{Cost statistics for one run over all tasks under the same budgets as in \cref{tab:main}. Top: number of LLM calls and total raw tokens~(input + output, in millions) for Best-of-N and \codename. Bottom: breakdown of \codename's LLM calls by operation, along with the number of embedding calls used for retrieval.}
\label{tab:cost_details}
\small
\scalebox{0.85}{
\begin{tabular}{@{}llccccc@{}}
\toprule
 & & \textbf{HMMT} & \textbf{BigCodeBench} & \textbf{LiveCodeBench} & \textbf{ScienceWorld} & \textbf{PDDL} \\
\midrule
\multirow{2}{*}{LLM calls} & Best-of-N & 2790 & 29748 & 2480 & -- & -- \\
 & \codename & 2398 & 22487 & 1474 & 3348 & 2017 \\
\midrule
\multirow{2}{*}{Raw tokens (M)} & Best-of-N & 4.2 & 37 & 5.4 & -- & -- \\
 & \codename & 6.5 & 49 & 4.9 & 9.6 & 6.5 \\
\midrule
\multicolumn{7}{@{}l}{\textit{\codename LLM call breakdown}} \\
 & Solution generation & 2080 & 15000 & 1200 & 1850 & 1098 \\
 & Scoring & 105 & 148 & 80 & 459 & 265 \\
 & Abstraction extraction & 208 & 3900 & 174 & 531 & 280 \\
 & Consolidation & 5 & 3439 & 20 & 508 & 374 \\
 & \textbf{Total LLM calls} & \textbf{2398} & \textbf{22487} & \textbf{1474} & \textbf{3348} & \textbf{2017} \\
\midrule
 & Embedding calls~(retrieval) & 563 & 8850 & 420 & 2592 & 1471 \\
\bottomrule
\end{tabular}
}
\end{table}

\begin{table}[t]
\centering
\caption{Agreement between the self-evaluation signal and the ground-truth evaluation on each benchmark. Precision is the fraction of self-accepted solutions that truly pass; recall is the fraction of truly passing solutions that are self-accepted. The overestimate rate is the fraction of solutions that are self-accepted but truly fail; the underestimate rate is the fraction that are self-rejected but truly pass.}
\label{tab:self_eval}
\small
\begin{tabular}{@{}lccccc@{}}
\toprule
 & \textbf{HMMT} & \textbf{BigCodeBench} & \textbf{LiveCodeBench} & \textbf{ScienceWorld} & \textbf{PDDL} \\
\midrule
ROC-AUC & 89.6 & 67.3 & 79.5 & 72.3 & 99.3 \\
Precision & 86.8 & 43.8 & 80.3 & 30.0 & 100.0 \\
Recall & 90.1 & 22.9 & 60.7 & 61.2 & 97.8 \\
F1 & 88.4 & 30.1 & 69.1 & 40.3 & 98.9 \\
\midrule
Overestimate rate & 8.2 & 9.2 & 8.4 & 16.1 & 0.0 \\
Underestimate rate & 5.9 & 24.0 & 22.1 & 4.4 & 0.4 \\
\bottomrule
\end{tabular}
\end{table}

\begin{figure}[t]
    \centering
    \includegraphics[width=.7\linewidth]{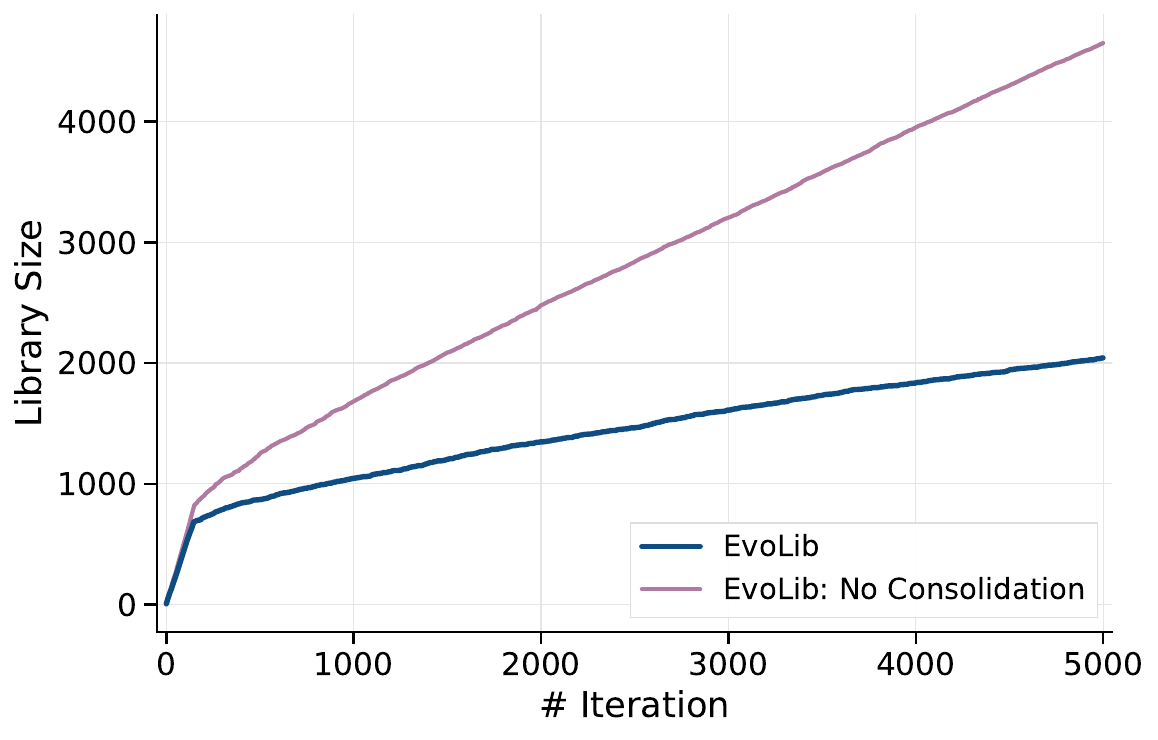}
\caption{Growth of the library size over iterations with and without consolidation on BigCodeBench. The x-axis shows the number of iterations and the y-axis shows the number of abstractions stored in the library.}
\label{fig:lib_size_curve}
\end{figure}

\begin{figure}[t]
    \begin{subfigure}[b]{0.5\textwidth}
        \includegraphics[width=\textwidth]{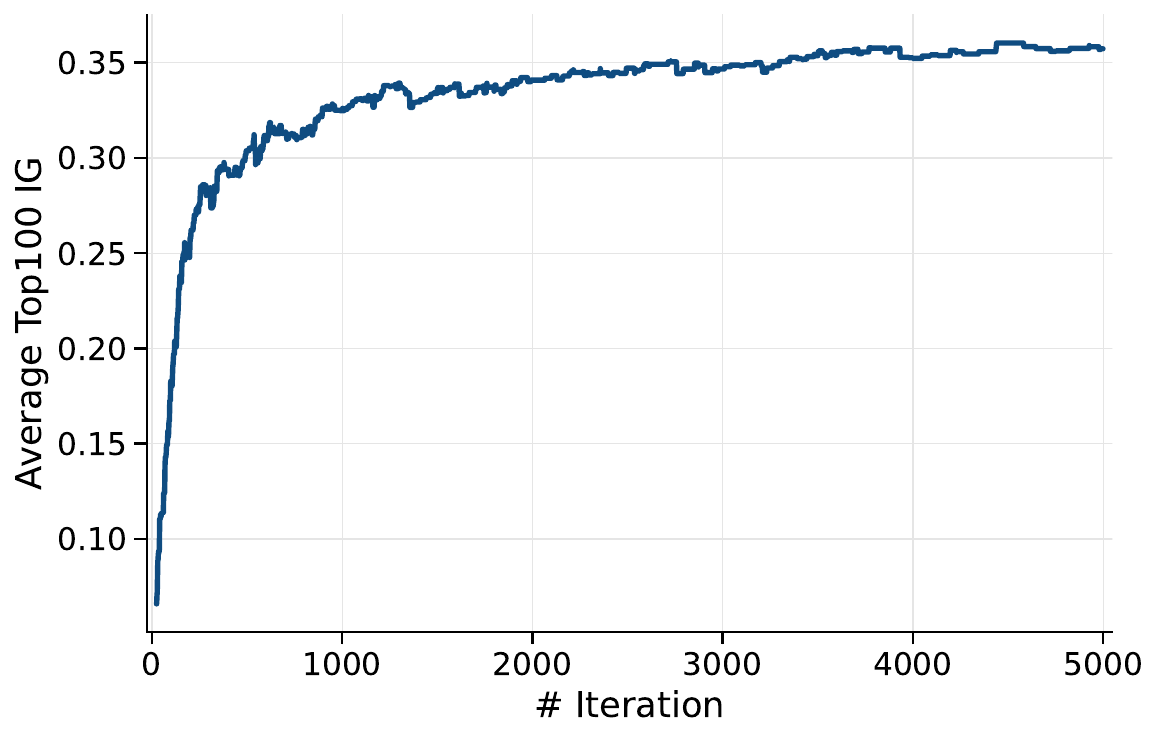}
        \caption{Average IG Among Top 100 Entries}
    \end{subfigure}
    \begin{subfigure}[b]{0.5\textwidth}
        \includegraphics[width=\textwidth]{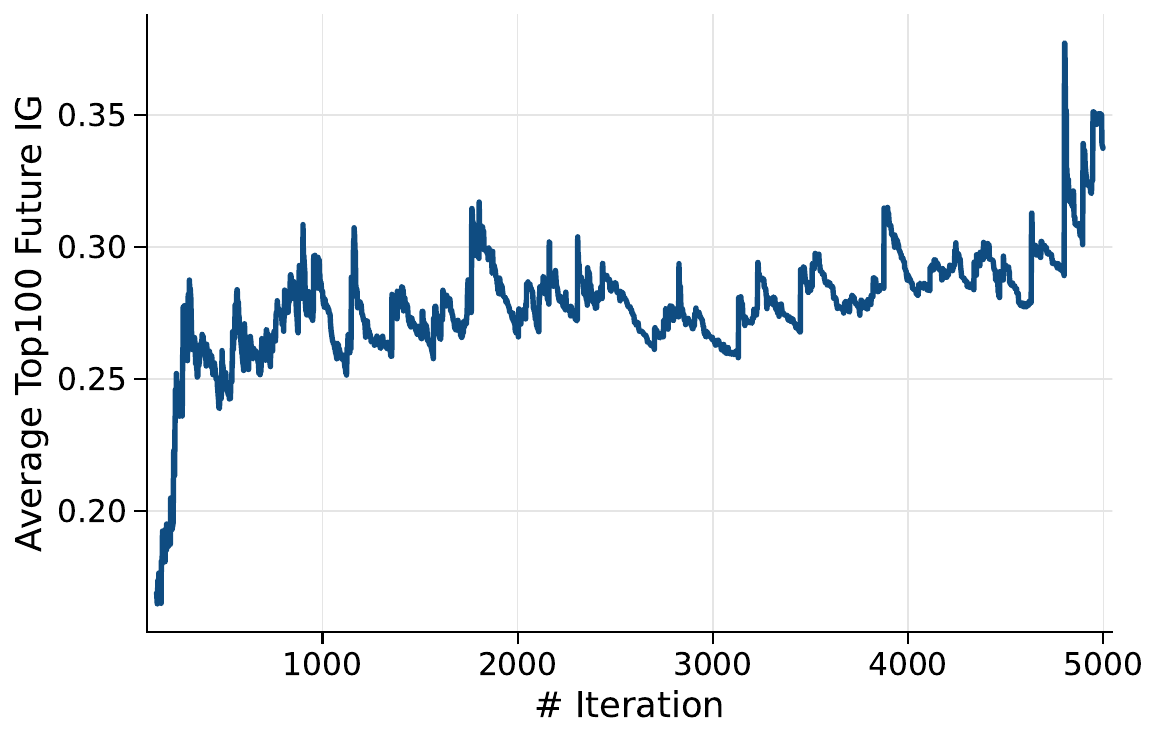}
        \caption{Average Future IG Among Top 100 Entries}
    \end{subfigure}
\caption{Average Information Gain (IG) and Future Information Gain (Future IG) among the top 100 abstractions in the library over iterations on BigCodeBench. The x-axis shows the number of iterations and the y-axis shows the average score for the top-ranked abstractions.}
\label{fig:avg_top_weight}
\end{figure}

\begin{table*}[t]
\centering
\footnotesize
\setlength{\tabcolsep}{4pt}
\begin{tabular}{p{0.03\linewidth}p{0.95\linewidth}}
\toprule
\multicolumn{2}{c}{\textbf{Consolidation Case Study}} \\
\midrule

\textbf{1} & \textbf{(Over-generalized $\rightarrow$ Grounded)} \\

& \textbf{Old Insight:} \\
& If you need to move multiple objects between two rooms and have two free grippers, then pick up two different objects in the source room before moving, so you can carry both in one trip. \\

& \textbf{New Insight:} \\
& If both grippers are free and there are objects in the same room that need relocation, then use both grippers simultaneously to pick objects, reducing the number of move--drop cycles. \\

& \textbf{Consolidated Insight:} \\
& If both grippers are free and there are multiple objects in the same room that need relocation, then pick up two different objects simultaneously before moving, so you can carry both in one trip and reduce the number of move--drop cycles. \\

\midrule

\textbf{2} & \textbf{(Over-generalized $\rightarrow$ Grounded)} \\

& \textbf{Old Insight:} \\
& If a calculated metric such as AUC or any other value can result in NaN due to invalid inputs or edge cases, then handle such cases explicitly (e.g., skipping computation, default values, or raising errors). \\

& \textbf{New Insight:} \\
& If a metric (e.g., AUC) can result in NaN due to edge cases (e.g., all true labels are the same $\rightarrow$ undefined ROC), then explicitly detect and handle this condition. \\

& \textbf{Consolidated Insight:} \\
& If a calculated metric such as AUC or any other value can result in NaN due to invalid inputs or edge cases, such as all true labels being the same leading to an undefined ROC curve, then handle such cases explicitly (e.g., skipping computation, default values, or raising meaningful errors). \\
\midrule

\textbf{3} & \textbf{(Overfitting $\rightarrow$ General)} \\

& \textbf{Old Skill:} \\
& \texttt{def validate\_dataframe\_columns(data, min\_columns):} \\
& \texttt{\ \ if not isinstance(data, pd.DataFrame):} \\
& \texttt{\ \ \ \ raise ValueError(...)} \\
& \texttt{\ \ if data.shape[1] $<$ min\_columns:} \\
& \texttt{\ \ \ \ raise ValueError(...)} \\

& \textbf{New Skill:} \\
& \texttt{def validate\_minimum\_categories(data, columns, min\_categories=2):} \\
& \texttt{\ \ for col in columns:} \\
& \texttt{\ \ \ \ if data[col].nunique() $<$ min\_categories:} \\
& \texttt{\ \ \ \ \ \ raise ValueError(...)} \\

& \textbf{Consolidated Skill:} \\
& \texttt{def validate\_dataframe(data, min\_columns=None, column\_category\_constraints=None):} \\
& \texttt{\ \ if not isinstance(data, pd.DataFrame):} \\
& \texttt{\ \ \ \ raise ValueError(...)} \\
& \texttt{\ \ if min\_columns is not None:} \\
& \texttt{\ \ \ \ if data.shape[1] $<$ min\_columns:} \\
& \texttt{\ \ \ \ \ \ raise ValueError(...)} \\
& \texttt{\ \ if column\_category\_constraints is not None:} \\
& \texttt{\ \ \ \ for col, k in column\_category\_constraints.items():} \\
& \texttt{\ \ \ \ \ \ if col not in data.columns:} \\
& \texttt{\ \ \ \ \ \ \ \ raise ValueError(...)} \\
& \texttt{\ \ \ \ \ \ if data[col].nunique() $<$ k:} \\
& \texttt{\ \ \ \ \ \ \ \ raise ValueError(...)} \\

\bottomrule
\end{tabular}
\caption{Consolidation cases illustrating generalization and grounding effects in \codename.}
\label{tab:consolidation_cases}
\end{table*}

\begin{table}[t]
\centering
\caption{Success and Progress Rate on PDDL under different task orders. We report mean $\pm$ standard deviation over three runs for each method.}
\label{tab:pddl_order}
\small
\begin{tabular}{@{}lcccccc@{}}
\toprule
\multirow{2}{*}{\textbf{Method}}
& \multicolumn{2}{c}{\textbf{Easy $\rightarrow$ Hard}}
& \multicolumn{2}{c}{\textbf{Hard $\rightarrow$ Easy}}
& \multicolumn{2}{c}{\textbf{Random}} \\
\cmidrule(lr){2-3} \cmidrule(lr){4-5} \cmidrule(lr){6-7}
& Success & Progress & Success & Progress & Success & Progress \\
\midrule
DC
& 66.7 $\pm$ 2.9 & 82.2 $\pm$ 2.5
& 65.6 $\pm$ 1.9 & 80.7 $\pm$ 0.6
& 65.6 $\pm$ 1.0 & 81.5 $\pm$ 0.8 \\
\codename
& \textbf{71.1 $\pm$ 2.5} & \textbf{85.5 $\pm$ 3.0}
& \textbf{68.9 $\pm$ 4.8} & \textbf{86.3 $\pm$ 0.7}
& \textbf{74.4 $\pm$ 1.9} & \textbf{88.1 $\pm$ 1.5} \\
\bottomrule
\end{tabular}
\end{table}

\paragraph{Cost-Performance Curve on HMMT.}
\cref{fig:main_cost_performance_hmmt} shows the cost–performance curve on HMMT.
\codename maintains a consistent advantage across the compute range, similar to the trends observed on coding benchmarks in the main paper. Although \codename adds an average of~1-2 LLM calls per iteration for scoring, maintaining and updating the library, it improves performance more rapidly with fewer attempts than the TTS approaches.

\paragraph{Cost breakdown.}
\label{sec:cost_details}
\cref{tab:cost_details} reports the number of LLM calls, raw token counts, and a breakdown of \codename's LLM calls by operation for one run over all tasks under the same budgets as in \cref{tab:main}. Over half of \codename's LLM calls~(54--87\%) are used for solution generation, while the remaining calls are used for scoring, abstraction extraction, and consolidation. Retrieval additionally requires embedding calls, which are substantially cheaper than generative LLM calls. 

\paragraph{Quality of self-evaluation.}
\label{sec:self_eval_details}
\cref{tab:self_eval} compares the self-evaluation signal used by \codename with the ground-truth evaluation on each benchmark. The quality of the self-evaluation signal varies across benchmarks. It is especially high on PDDL~(ROC-AUC 99.3), where the environment responses contain rich information for the model to accurately judge task progress. On the coding benchmarks, the signal is considerably noisier~(ROC-AUC 67.3 on BigCodeBench and 79.5 on LiveCodeBench), and the synthetic tests reject many truly correct solutions~(underestimate rates of 24.0\% and 22.1\%). Even so, \codename still leads to substantial gains on these benchmarks and improves performance more efficiently than the baselines as test-time compute increases~(\cref{tab:main,fig:main_cost_performance}).

\paragraph{Weaker self-evaluation model.}
We further experiment with a weaker judge on BigCodeBench, using Qwen3.5-9B to generate the synthetic tests while using GPT-4o for solution generation and library-related operations. In this setting, \codename achieves a pass rate of 35.4\%, still outperforming Best-of-N under the same setting by~1.6 points~(33.8\%).

\paragraph{Library growth and the role of consolidation.}
\cref{fig:lib_size_curve} compares the growth of the library with and without consolidation.
Without consolidation, the number of entries increases nearly linearly with the number of iterations, reflecting the accumulation of task-specific abstractions.
In contrast, with consolidation, the library grows quickly at first and then much more slowly, staying under 2000 entries after 5000 iterations, compared with over 4000 entries without consolidation.
Case 3 in \cref{tab:consolidation_cases} provides a concrete example of the consolidation process.
A newly extracted function and an existing function are merged into a more general abstraction that subsumes both behaviors.
This illustrates how the library evolves from specific, task-level functions into abstractions that apply across multiple problems.
Such generalization is central to enabling transfer across tasks without explicit supervision.
These results support the role of consolidation in controlling memory growth while encouraging the emergence of generic abstractions.

\paragraph{Evolution of abstraction quality.}
\cref{fig:avg_top_weight} examines how the quality of the most useful abstractions evolves over time.
Both the average IG and average Future IG among the top entries increase as the number of iterations grows.
The increase in IG indicates that the library progressively contains abstractions that directly improve solution quality.
The increase in Future IG suggests that the library also becomes better at generating abstractions that enable further learning.
Together, these trends provide evidence that the weighting mechanism prioritizes abstractions that contribute both immediate and future utility.

\paragraph{Effect of task order in continual learning.}
\label{paragraph:task_order}
\cref{tab:pddl_order} compares \codename with Dynamic Cheatsheet~(DC) under different task orders, including easy-to-hard, hard-to-easy, and random permutations.
\codename consistently outperforms DC across all settings, achieving higher success rates and progress rates.
Notably, the advantage is largest under random ordering, where no curriculum structure is present, suggesting that \codename is more robust to variations in the task stream.

Importantly, since PDDL comprises four different types of tasks that require different types of actions and planning strategies, strong performance under random ordering indicates that \codename can effectively handle and continually learn from a heterogeneous mixture of task types without relying on a structured curriculum.
This property is desirable in real-world scenarios, where an agent must respond to diverse user requests in arbitrary order and still accumulate useful abstractions over time.

Overall, these results support the hypothesis that \codename enables continual learning that is less dependent on task ordering, whereas methods based on linear knowledge updates are more sensitive to the ordering.

\begin{table*}[t]
\centering
\footnotesize
\setlength{\tabcolsep}{4pt}
\begin{tabular}{p{0.02\linewidth}p{0.96\linewidth}}
\toprule
\multicolumn{2}{c}{\textbf{Top Insights (BigCodeBench)}} \\
\midrule
1 & If (your function or program processes input that requires specific formatting or encoding, such as hexadecimal strings or UTF-8 encoded data), then (handle potential decoding errors gracefully by implementing checks and providing meaningful error messages before proceeding with further processing). \\
\midrule
2 & If a function depends on external systems (e.g., network, file I/O, or hardware) to produce results, then ensure to mock or simulate these external dependencies during testing to avoid false negatives caused by environmental factors or unhandled edge cases. \\
\midrule
3 & If (a function interacts with a directory or file system), then consider the case (the directory or file may not exist) and handle it explicitly by checking for existence before performing operations, or by using try-except blocks to catch and handle \texttt{FileNotFoundError} or similar exceptions. \\
\midrule
4 & If (a function processes external files or data inputs), then ensure to validate the existence, format, and content of the files or data before proceeding with further operations. Specifically, check for conditions such as file existence, non-empty content, and adherence to expected structure or format. Handle errors gracefully by providing meaningful error messages, consider edge cases like empty files or missing data to prevent runtime exceptions, raise appropriate exceptions or provide fallback mechanisms for invalid data, and explicitly handle cases where the input might be empty, corrupted, or invalid. \\
\midrule
5 & If your code involves calculating metrics like AUC or other statistical measures, then ensure that the input data (e.g., true labels and predicted probabilities) is valid and does not contain edge cases such as all true labels being the same class (e.g., all 0s or all 1s) or predicted probabilities being constant, as these can result in undefined or NaN values. Consider adding checks to validate the input data and handle such cases explicitly, such as skipping the calculation, providing a meaningful default value, or raising an informative error. \\
\midrule
\multicolumn{2}{c}{\textbf{Top Skills (BigCodeBench)}} \\
\midrule
1 & \texttt{def validate\_range(range\_low, range\_high):} \\
  & \texttt{""" Validate that the lower bound is less than the upper bound. Parameters: range\_low (int): Lower bound of the range. range\_high (int): Upper bound of the range. Raises: ValueError: If range\_low is not less than range\_high. """} \\
\midrule
2 & \texttt{def fetch\_and\_prepare\_boston\_data(data\_url):} \\
  & \texttt{""" Fetches and prepares the Boston Housing dataset from the given URL. Parameters: data\_url (str): URL to the Boston Housing dataset. Returns: pd.DataFrame: A DataFrame containing the prepared Boston Housing data. """} \\
\midrule
3 & \texttt{def task\_func(kwargs, target\_dir="non\_none\_files"):} \\
  & \texttt{""" Process files from a dictionary by checking if the file exists, and if it has content, then copies it to a target directory. Parameters: kwargs (dict): dictionary of file paths to content. target\_dir (str): destination directory. Returns: list of copied file paths. """} \\
\midrule
4 & \texttt{def task\_func(df, target\_column):} \\
  & \texttt{""" Train a random forest classifier to perform classification and plot feature importances. The plot uses feature importance scores on the x-axis and feature names on the y-axis, sorted in descending order. Returns the trained model and plot Axes. """} \\
\midrule
5 & \texttt{def encrypt\_aes\_key\_with\_rsa(aes\_key, public\_key):} \\
  & \texttt{""" Encrypts an AES key using an RSA public key. Parameters: aes\_key (bytes): The AES key to encrypt. public\_key (rsa.PublicKey): The RSA public key. Returns: bytes: The encrypted AES key. """} \\
\bottomrule
\end{tabular}
\caption{Top insights and skills with high Future IG in \codename on BigCodeBench.}
\label{tab:evolib_bigcode_top_abstractions}
\end{table*}

\begin{table*}[t]
\centering
\footnotesize
\setlength{\tabcolsep}{4pt}
\begin{tabular}{p{0.02\linewidth}p{0.96\linewidth}}
\toprule
\multicolumn{2}{c}{\textbf{Top Insights (LiveCodeBench)}} \\
\midrule

1 & If you’re maintaining counts of available items by splitting them into “below” and “above” a moving threshold, then after picking an item you must decrement the count of whichever subgroup it came from (or adjust your pointer), not just the total used count, so your future availability calculations remain correct. \\

\midrule
2 & If you define a block in Python (such as a function, class, loop, or conditional) ending with a colon, then do include at least one indented statement (even just a \texttt{pass}) immediately after, otherwise you’ll get an "expected an indented block" syntax error. \\

\midrule
3 & If you split your digit‐DP into two passes (one for numbers that contain a special digit and one for numbers that don’t), then do not treat every zero the same—use a “started” flag so that leading zeros aren’t counted as actual zeros, only set your “have\_zero” flag when you see a zero after you’ve “started,” and in the zero‐free pass forbid zeros once “started.” This way your two DPs form a disjoint, exhaustive partition and you won’t undercount or overcount any numbers. \\

\midrule
4 & If your problem involves multiple agents that cannot occupy the same vertex at the same time—but may visit the same vertex at different times—then do not model these conflicts as permanent vertex‐capacity constraints. Instead, you must encode the temporal dimension (for example by building a time‐expanded graph or by BFS on the joint (position,time) state). \\

\midrule
5 & If you’re using binary search to maximize or minimize an integer parameter via a feasibility predicate, then verify that your initial bounds cover the entire range of possible values, ensure the predicate is monotonic, and test edge cases; otherwise you may cut off the real optimum or get the wrong threshold. \\

\midrule
\multicolumn{2}{c}{\textbf{Top Skills (LiveCodeBench)}} \\
\midrule

1 & \texttt{def greedy\_interval\_cover(sorted\_intervals, cover\_L, cover\_R):} \\
  & \texttt{""" Greedy interval covering algorithm. Given intervals sorted by left endpoint, selects a minimum-cardinality sub-collection that covers [cover\_L, cover\_R]. Returns selected indices or empty list if impossible. """} \\

\midrule
2 & \texttt{def divisors\_from\_prime\_counts(prime\_counts: dict) -> list:} \\
  & \texttt{""" Generate all divisors of a number from its prime factorization using DFS over prime powers. """} \\

\midrule
3 & \texttt{def kmp\_search(text, pattern):} \\
  & \texttt{""" Return True if pattern occurs in text using the KMP algorithm. Time complexity: O(len(text) + len(pattern)). """} \\

\midrule
4 & \texttt{def tarjan\_scc(graph):} \\
  & \texttt{""" Compute strongly connected components of a directed graph using Tarjan's algorithm. Returns a list of components. """} \\

\midrule
5 & \texttt{def binary\_search\_last\_true(lo, hi, pred):} \\
  & \texttt{""" Given a monotonic predicate pred(x), find the maximum x in [lo, hi] such that pred(x) is True. """} \\

\bottomrule
\end{tabular}
\caption{Top insights and skills with high Future IG in \codename on LiveCodeBench.}
\label{tab:evolib_livecode_top_abstractions}
\end{table*}

\begin{table*}[t]
\centering
\footnotesize
\setlength{\tabcolsep}{4pt}
\begin{tabular}{p{0.02\linewidth}p{0.96\linewidth}}
\toprule
\multicolumn{2}{c}{\textbf{Top Insights (HMMT)}} \\
\midrule

1 & If a single oriented half-space on two parallel convex polygons selects “sup–blocks” of vertices, then those blocks on the two polygons must share the same extremal index (they start at the same vertex) unless one of the blocks is empty or the entire polygon. \\ \midrule

2 & If you use inclusion–exclusion to count “bad” subsets defined by forbidden substructures (cuts, patterns, etc.), you must list and include all minimal forbidden substructures—in particular those of higher cardinality that do not contain smaller ones—to avoid undercounting. \\ \midrule

3 & If a point induces two triangles with a common “angle-ratio” condition and known side-length ratio, then view the situation as a spiral similarity centered at that point—write its rotation angle and scale factor explicitly, and impose the “sum of the four angles around a point equals 360°” to link the small triangle’s base angle to the given interior angle. \\ \midrule

4 & If you use the Law of Sines in two triangles sharing no common angle, then a ratio of side-lengths such as BP/PC generally equals a product of three sine-ratios, not simply sin x/sin y. Always carry along the extra factors from both sine laws. \\ \midrule

5 & If you need the sign of sin(x) by comparing x to multiples of $\pi$, then use floor(x/$\pi$) mod 2: sin(x)$>$0 exactly when floor(x/$\pi$) is even. Moreover, parity (floor($2^n\cdot\alpha$)) equals the nth binary digit of $\alpha$. \\

\midrule
\multicolumn{2}{c}{\textbf{Top Skills (HMMT)}} \\
\midrule

1 & \textbf{Eliminate variables by combining constraints:} When you have two constraints (e.g., a right-angle/power relation plus a circle/quadratic relation), substitute one into the other to eliminate a squared term and derive a simpler (often linear) relationship among the remaining variables. \\ \midrule
2 & \textbf{Exploit symmetry to reuse a count:} If a configuration has an involutive symmetry (swapping roles, reversing a half-space, etc.), map the “opposite case” to the original one to conclude both cases contribute equally, rather than recounting from scratch. \\ \midrule
3 & \textbf{Re-encode divisibility as a product order:} For sets defined by prime-power structure, represent elements by exponent vectors; then divisibility becomes coordinatewise comparison, turning the problem into reasoning about a product of chains (a grid poset). \\ \midrule
4 & \textbf{Parameterize points on a segment affinely:} To locate points at specified distances along a segment, express each point as an affine combination $P = (1-t)B + tC$ with $t$ equal to the normalized arc-length fraction; then read off coordinates or derived quantities cleanly. \\ \midrule
5 & \textbf{Reduce interacting motion to a standard stochastic process:} When multiple moving objects merge upon contact, reinterpret the dynamics as coalescing random walks (or an equivalent Markov process) on an appropriate state space, enabling use of known results or simpler expectation calculations. \\

\bottomrule
\end{tabular}
\caption{Top insights and skills with high Future IG in \codename on HMMT.}
\label{tab:evolib_hmmt_top_abstractions}
\end{table*}

\begin{table*}[t]
\centering
\footnotesize
\setlength{\tabcolsep}{4pt}
\begin{tabular}{p{0.02\linewidth}p{0.96\linewidth}}
\toprule
\multicolumn{2}{c}{\textbf{Top Insights (ScienceWorld)}} \\
\midrule
1 & If you are navigating to a known target location, then first confirm the correct connecting doorway by using “look around” in your current room, plan the most direct route, and avoid back-and-forth movements that don’t add information. \\ \midrule
2 & If you need to choose among multiple objects, first list or note all candidates, then compare their key attributes (e.g. life span) before focusing on one. \\ \midrule
3 & If you intend to select from an ambiguity menu, do not type just the index; you must include the action keyword plus the exact indexed object name. \\ \midrule
4 & If you have heated an ingredient and see “Progress 0.5/1.0,” then check with the appropriate measuring device to complete the task. \\ \midrule
5 & If a substance must reach a phase change, then ensure the heating source is fully activated and give sufficient waiting time (or confirm rising temperature) before checking for melting. \\

\midrule
\multicolumn{2}{c}{\textbf{Top Skills (ScienceWorld)}} \\
\midrule
1 & \textbf{Functionality:} Heat a substance in a container using the stove until it reaches or exceeds a target temperature. \\
  & \textbf{Condition:} When you need to raise the temperature of whatever is in the vessel to a desired threshold. \\
  & \textbf{Input Variables:} Stove, Thermometer, Vessel, TargetTemp. \\
  & \textbf{Steps:} \\
  & \quad \texttt{move Vessel to Stove} \\
  & \quad \texttt{activate Stove} \\
  & \quad \texttt{wait1} \\
  & \quad \texttt{wait1} \\
  & \quad \texttt{wait1} \\
  & \quad \texttt{use Thermometer on Vessel} \\
  & \quad \texttt{(if temperature < TargetTemp, repeat wait and measure)} \\
  & \quad \texttt{deactivate Stove} \\
\midrule

2 & \textbf{Functionality:} Deliver or deposit an object from your inventory into a target container. \\
  & \textbf{Condition:} When you have the object in inventory and need to place it into a named container. \\
  & \textbf{Input Variables:} OBJ, DEST\_LOC, CONTAINER. \\
  & \textbf{Steps:} \\
  & \quad \texttt{(if not at DEST\_LOC) go to DEST\_LOC} \\
  & \quad \texttt{look around} \\
  & \quad \texttt{move OBJ to CONTAINER} \\
\midrule

3 & \textbf{Functionality:} Retrieve an item from a storage container. \\
  & \textbf{Condition:} When an object needed for a task is inside a container. \\
  & \textbf{Input Variables:} OBJ, CONTAINER. \\
  & \textbf{Steps:} \\
  & \quad \texttt{open CONTAINER} \\
  & \quad \texttt{look in CONTAINER} \\
  & \quad \texttt{pick up OBJ} \\
\midrule

4 & \textbf{Functionality:} Verify the physical state of a substance after a process. \\
  & \textbf{Condition:} After performing a process intended to change a substance’s state (e.g., heating or freezing). \\
  & \textbf{Input Variables:} STORAGE, CONTAINER, SUBSTANCE. \\
  & \textbf{Steps:} \\
  & \quad \texttt{(if STORAGE exists) open STORAGE} \\
  & \quad \texttt{open CONTAINER} \\
  & \quad \texttt{look in CONTAINER} \\
  & \quad \texttt{focus on SUBSTANCE} \\
  & \quad \texttt{look at SUBSTANCE} \\
\midrule

5 & \textbf{Functionality:} Wait until an object changes state over time. \\
  & \textbf{Condition:} When an environmental or physical process requires time to complete. \\
  & \textbf{Input Variables:} desired\_state. \\
  & \textbf{Steps:} \\
  & \quad \texttt{wait} \\
  & \quad \texttt{(or) wait1} \\
  & \quad \texttt{look around} \\
  & \quad \texttt{repeat until state changes} \\

\bottomrule
\end{tabular}
\caption{Top insights and skills with high Future IG in \codename on ScienceWorld.}
\label{tab:evolib_scienceworld_top_abstractions}
\end{table*}

\begin{table*}[t]
\centering
\footnotesize
\setlength{\tabcolsep}{4pt}
\begin{tabular}{p{0.02\linewidth}p{0.96\linewidth}}
\toprule
\multicolumn{2}{c}{\textbf{Top Insights (PDDL)}} \\
\midrule
1 & If working with nuts on a hub, then jack up the hub before undoing or removing the wheel; after wheel removal or installation, refasten the nuts and jack down; and ensure the hub is on the ground before tightening any loose nut. \\ \midrule
2 & If you leave a shot glass on the table after cleaning, then you can pour into it directly from a shaker without grasping it, as long as you satisfy the hand-holding requirements of the pour action. \\ \midrule
3 & If you finish dropping the last carried object, then check immediately for any remaining objects in the other room to avoid idle moves. \\ \midrule
4 & If the hub is on the ground, then you must loosen the nuts before jacking up; otherwise removal is impossible. \\ \midrule
5 & If a block you want is not on the table, then unstack it from whichever block it’s on before attempting to pick it up. \\

\midrule
\multicolumn{2}{c}{\textbf{Top Skills (PDDL)}} \\
\midrule
1 & \textbf{Functionality:} Load two specified ingredients into a clean shaker using two separate shot glasses. \\
  & \textbf{Condition:} The shaker is clean and empty; both shot glasses are clean and empty; both hands are empty; dispensers are available. \\
  & \textbf{Input Variables:} shot1, ingredient1, shot2, ingredient2, hand, other\_hand, shaker, dispenser1, dispenser2, l0, l1, l2. \\
  & \textbf{Steps:} \\
  & \quad \texttt{hand grasp shot1} \\
  & \quad \texttt{fill-shot shot1 ingredient1 hand other\_hand dispenser1} \\
  & \quad \texttt{pour-shot-to-clean-shaker shot1 ingredient1 shaker hand l0 l1} \\
  & \quad \texttt{clean-shot shot1 ingredient1 hand other\_hand} \\
  & \quad \texttt{fill-shot shot2 ingredient2 hand other\_hand dispenser2} \\
  & \quad \texttt{pour-shot-to-used-shaker shot2 ingredient2 shaker hand l1 l2} \\
  & \quad \texttt{hand leave shot2} \\
\midrule

2 & \textbf{Functionality:} Remove a wheel from a hub and stow it in the boot. \\
  & \textbf{Condition:} The hub is on the ground, the nut is tight, tools are available, and the boot is open. \\
  & \textbf{Input Variables:} hub, wheel, nut, boot. \\
  & \textbf{Steps:} \\
  & \quad \texttt{loosen nut hub} \\
  & \quad \texttt{jack-up hub} \\
  & \quad \texttt{undo nut hub} \\
  & \quad \texttt{remove-wheel wheel hub} \\
  & \quad \texttt{put-away wheel boot} \\
  & \quad \texttt{jack-down hub} \\
\midrule

3 & \textbf{Functionality:} Fetch the basic tools (wrench and jack) from a closed container. \\
  & \textbf{Condition:} The container is closed but unlocked, and tools are inside. \\
  & \textbf{Input Variables:} container, wrench, jack. \\
  & \textbf{Steps:} \\
  & \quad \texttt{open container} \\
  & \quad \texttt{fetch wrench container} \\
  & \quad \texttt{fetch jack container} \\
\midrule

4 & \textbf{Functionality:} Move a block from one block to another. \\
  & \textbf{Condition:} The arm is empty; block X is clear and on block U; block V is clear. \\
  & \textbf{Input Variables:} X, U, V. \\
  & \textbf{Steps:} \\
  & \quad \texttt{unstack X U} \\
  & \quad \texttt{stack X V} \\
\midrule

5 & \textbf{Functionality:} Shake a two-ingredient mixture in a shaker to form a cocktail. \\
  & \textbf{Condition:} The shaker contains the correct ingredients and is unshaken. \\
  & \textbf{Input Variables:} cocktail, ingredient1, ingredient2, shaker, hand1, hand2. \\
  & \textbf{Steps:} \\
  & \quad \texttt{hand1 grasp shaker} \\
  & \quad \texttt{shake cocktail ingredient1 ingredient2 shaker hand1 hand2} \\

\bottomrule
\end{tabular}
\caption{Top insights and skills with high Future IG in \codename on PDDL.}
\label{tab:evolib_pddl_top_abstractions}
\end{table*}

\paragraph{\codename addresses the failure modes in continuously updated abstract memory.}
Recent work shows that continuously updated abstract memory in existing approaches can degrade performance through three major failure modes: \emph{misgrouping}, where experiences from different types of tasks are consolidated into the same abstraction; \emph{interference}, where repeated rewriting strips away applicability conditions and turns useful lessons into overly broad guidance; and \emph{overfit}, where memory becomes tied to narrow surface patterns rather than reusable strategies~\cite{zhang2026usefulmemoriesfaultycontinuously}.
For example, Dynamic Cheatsheet relies solely on the LLM to repeatedly rewrite the whole memory, which can lead to misgrouping and interference. In contrast, \codename extracts knowledge from a single problem and trajectory at a time, consolidates only highly related abstractions, and propagates credit along evolution chains, so that specialized abstractions can coexist and evolve nonlinearly.
Our experiments show that \codename avoids or substantially alleviates these issues:

\textbf{Misgrouping.} \codename is less susceptible to misgrouping because its consolidation is localized rather than global: two abstractions are merged only when they are close in the embedding space and are judged semantically and functionally similar. \cref{tab:pddl_order} provides further evidence: \codename remains strong even under random task ordering in PDDL, where four task types are interleaved, suggesting that the library can accumulate useful abstractions from a heterogeneous stream without requiring a clean curriculum to keep different task families separated.

\textbf{Interference.} The interference failure mode arises when repeated rewriting and consolidation remove the conditions under which a lesson is valid, causing an abstraction learned from one setting to mislead behavior in another. \codename mitigates this effect as it extracts an abstraction from one trajectory at a time and only consolidates it with another abstraction when they are semantically and functionally similar. Case~1 and~2 in \cref{tab:consolidation_cases} show that the consolidation process in \codename can move abstractions in the opposite direction: it adds more grounded conditions to the insight statements rather than smoothing them away into generic advice.
In addition, an abstraction that interferes with a broad range of tasks will receive low Future IG scores, decreasing its chance to be sampled during inference. As shown in \cref{tab:evolib_bigcode_top_abstractions,tab:evolib_livecode_top_abstractions,tab:evolib_hmmt_top_abstractions,tab:evolib_pddl_top_abstractions,tab:evolib_scienceworld_top_abstractions}, the top high-Future-IG abstractions remain concrete and applicable to only a subset of tasks.

\textbf{Overfit.} The overfit failure mode occurs when memory tracks recurring surface regularities of past instances instead of extracting a strategy that transfers to related tasks.
This problem is mitigated in \codename by the consolidation and weighting mechanism. An abstraction that is specific to a certain task can be consolidated with a new one~(from a related but different task) into a more general abstraction, as shown by Case~3 in \cref{tab:consolidation_cases}. Furthermore, abstractions that remain overly specific will then receive low Future IG scores and thus low sampling weights. As shown in \cref{tab:evolib_bigcode_top_abstractions,tab:evolib_livecode_top_abstractions,tab:evolib_hmmt_top_abstractions,tab:evolib_pddl_top_abstractions,tab:evolib_scienceworld_top_abstractions}, the highest-Future-IG entries are typically reusable procedures, checks, or reasoning templates that are applicable to similar tasks.

\section*{NeurIPS Paper Checklist}
\begin{enumerate}

\item {\bf Claims}
    \item[] Question: Do the main claims made in the abstract and introduction accurately reflect the paper's contributions and scope?
    \item[] Answer: \answerYes{}
    \item[] Justification: The abstract and introduction clearly state the main contribution (EvoLib) and its scope, including knowledge accumulation without parameter updates and empirical improvements across benchmarks (Sections 1 and Abstract).

\item {\bf Limitations}
    \item[] Question: Does the paper discuss the limitations of the work performed by the authors?
    \item[] Answer: \answerYes{}
    \item[] Justification: The paper includes a dedicated discussion of limitations (Section 5, “Limitations”).

\item {\bf Theory assumptions and proofs}
    \item[] Question: For each theoretical result, does the paper provide the full set of assumptions and a complete (and correct) proof?
    \item[] Answer: \answerNA{}
    \item[] Justification: The paper does not include any theoretical results.

    \item {\bf Experimental result reproducibility}
    \item[] Question: Does the paper fully disclose all the information needed to reproduce the main experimental results of the paper to the extent that it affects the main claims and/or conclusions of the paper (regardless of whether the code and data are provided or not)?
    \item[] Answer: \answerYes{}
    \item[] Justification: The paper describes the algorithm, experimental setup, datasets, baselines, and evaluation metrics in sufficient detail to reproduce results, including pseudocode (Algorithm 1), method and dataset descriptions (Section 4 and Appendix A). The full code (including prompts) is released in the supplementary material.

\item {\bf Open access to data and code}
    \item[] Question: Does the paper provide open access to the data and code, with sufficient instructions to faithfully reproduce the main experimental results, as described in supplemental material?
    \item[] Answer: \answerYes{}
    \item[] Justification: The full code (including prompts) is released in the supplementary material. All datasets used are referenced.

\item {\bf Experimental setting/details}
    \item[] Question: Does the paper specify all the training and test details (e.g., data splits, hyperparameters, how they were chosen, type of optimizer) necessary to understand the results?
    \item[] Answer: \answerYes{}
    \item[] Justification: The paper specifies all the dataset, model, hyperparameter, evaluation, and configuration details in Sections 4 and Appendix A.

\item {\bf Experiment statistical significance}
    \item[] Question: Does the paper report error bars suitably and correctly defined or other appropriate information about the statistical significance of the experiments?
    \item[] Answer: \answerYes{}
    \item[] Justification: We report statistical significance of the main result in Table 1 and explain how they were calculated.

\item {\bf Experiments compute resources}
    \item[] Question: For each experiment, does the paper provide sufficient information on the computer resources (type of compute workers, memory, time of execution) needed to reproduce the experiments?
    \item[] Answer: \answerYes{}
    \item[] Justification: Figure 2 and B.1 shows the amount of compute needed (in token count) to reach varying levels of performance for each method on each dataset.
    
\item {\bf Code of ethics}
    \item[] Question: Does the research conducted in the paper conform, in every respect, with the NeurIPS Code of Ethics \url{https://neurips.cc/public/EthicsGuidelines}?
    \item[] Answer: \answerYes{}
    \item[] Justification: The research adheres to standard machine learning experimentation practices, uses public datasets, and does not involve human subjects, sensitive data or harmful applications as presented.

\item {\bf Broader impacts}
    \item[] Question: Does the paper discuss both potential positive societal impacts and negative societal impacts of the work performed?
    \item[] Answer: \answerYes{}
    \item[] Justification: Section 5 explicitly discusses this work as a stepping stone toward self-improving agents, which is a positive societal impact on effective AI assistant. The ``Limitations'' paragraph further discusses potential negative impact when there is a misalignment between human and LLMs' judgment.
    
\item {\bf Safeguards}
    \item[] Question: Does the paper describe safeguards that have been put in place for responsible release of data or models that have a high risk for misuse (e.g., pre-trained language models, image generators, or scraped datasets)?
    \item[] Answer: \answerNA{}
    \item[] Justification: The work does not release new high-risk models or datasets and does not involve potentially sensitive or misuse-prone assets.

\item {\bf Licenses for existing assets}
    \item[] Question: Are the creators or original owners of assets (e.g., code, data, models), used in the paper, properly credited and are the license and terms of use explicitly mentioned and properly respected?
    \item[] Answer: \answerYes{}
    \item[] Justification: All datasets, baseline methods, and models used are cited and used in accordance with their license.

\item {\bf New assets}
    \item[] Question: Are new assets introduced in the paper well documented and is the documentation provided alongside the assets?
    \item[] Answer: \answerYes{}
    \item[] Justification: The full code together with clear documentation are released in the supplementary material.

\item {\bf Crowdsourcing and research with human subjects}
    \item[] Question: For crowdsourcing experiments and research with human subjects, does the paper include the full text of instructions given to participants and screenshots, if applicable, as well as details about compensation (if any)? 
    \item[] Answer: \answerNA{}
    \item[] Justification: The research does not involve human subjects or crowdsourced data collection.

\item {\bf Institutional review board (IRB) approvals or equivalent for research with human subjects}
    \item[] Question: Does the paper describe potential risks incurred by study participants, whether such risks were disclosed to the subjects, and whether Institutional Review Board (IRB) approvals (or an equivalent approval/review based on the requirements of your country or institution) were obtained?
    \item[] Answer: \answerNA{}
    \item[] Justification: The study does not involve human participants and therefore does not require IRB approval.

\item {\bf Declaration of LLM usage}
    \item[] Question: Does the paper describe the usage of LLMs if it is an important, original, or non-standard component of the core methods in this research? Note that if the LLM is used only for writing, editing, or formatting purposes and does \emph{not} impact the core methodology, scientific rigor, or originality of the research, declaration is not required.
    \item[] Answer: \answerYes{}
    \item[] Justification: LLMs are a core component of the method (used for solution generation, evaluation, abstraction extraction, and consolidation), and their role is explicitly described throughout the paper (Sections 3 and 4).

\end{enumerate}

\end{document}